\documentclass[10pt, logo, onecolumn, copyright]{nv}

\usepackage{amsmath,amsfonts,bm}

\def\eqref#1{equation~\ref{#1}}

\def\1{\bm{1}}

\DeclareMathAlphabet{\mathsfit}{\encodingdefault}{\sfdefault}{m}{sl}
\SetMathAlphabet{\mathsfit}{bold}{\encodingdefault}{\sfdefault}{bx}{n}

\definecolor{linkc}{rgb}{0, 0.44, 0.74}
\definecolor{eqc}{rgb}{1, 0, 0}
\definecolor{newcitecolor}{rgb}{0,0.6,0}
\usepackage{url}
\usepackage{graphicx}
\usepackage{booktabs}
\usepackage{wrapfig}
\usepackage{amsmath}
\usepackage{xspace}
\usepackage{textgreek}
\usepackage{bm}
\usepackage{cleveref}
\usepackage{multirow}
\usepackage[dvipsnames]{xcolor}
\usepackage{colortbl}
\usepackage{orcidlink}
\usepackage{footmisc}
\usepackage{algorithm}
\usepackage{algpseudocode}
\usepackage{amsmath}
\usepackage{amsfonts}
\usepackage{listings}
\usepackage{graphicx}
\usepackage{caption}
\usepackage{float}
\usepackage{enumitem}
\usepackage{csquotes}
\usepackage{marvosym}
\usepackage{subcaption}

\definecolor{mygreen}{RGB}{34,139,34}
\definecolor{mylightblue}{RGB}{0,162,230}
\definecolor{deepyellow}{RGB}{255, 215, 0} 
\definecolor{catgray}{gray}{0.92}
\definecolor{nvidiagreen}{HTML}{76B900}

\makeatletter
\def\blfootnote#1{\xdef\@thefnmark{}\@footnotetext{\scriptsize #1}}
\makeatother

\hypersetup{
    colorlinks=true,
    urlcolor=nvidiagreen,
}

\newcommand{\ours}{SANA-Streaming\xspace}

\definecolor{assistantcolor}{HTML}{76B900}  
\newcommand{\rankfirst}[1]{\cellcolor{assistantcolor!32}#1}
\newcommand{\ranksecond}[1]{\cellcolor{assistantcolor!20}#1}

\title{\ours : Real-time Streaming Video Editing with Hybrid Diffusion Transformer}

\author{%
\vspace{-1.5em}
\centering
\fontsize{10pt}{18pt}\selectfont
Yuyang Zhao\textsuperscript{1$*$},
\quad Yicheng Pan\textsuperscript{3$*$},
\quad Qiyuan He\textsuperscript{1,4$*$},
\quad Jincheng Yu\textsuperscript{1$*$},
\quad Junsong Chen\textsuperscript{1,5$*$},
\quad Tian Ye\textsuperscript{1},
\quad Haozhe Liu\textsuperscript{1},
\quad Enze Xie\textsuperscript{1},
\quad Song Han\textsuperscript{1,2}
\\
\vspace{2.5mm}
{\normalsize \textsuperscript{1}NVIDIA ~~
\textsuperscript{2}MIT ~~
\textsuperscript{3}THU ~~
\textsuperscript{4}NUS ~~
\textsuperscript{5}HKU }
\\
\vspace{0.3em}
{\normalsize $^*$Equal contribution} \\
\vspace{0.3em}
{\normalsize Project Page: \textbf{\href{https://nvlabs.github.io/Sana/Streaming/}{https://nvlabs.github.io/Sana/Streaming}}}
\vspace{-2.5em}
}

\begin{document}

\begin{abstract}
Real-time streaming video-to-video editing (V2V) is critical for interactive applications such as live broadcasting and gaming, yet it remains a formidable challenge due to the stringent requirements for temporal consistency and inference throughput. 
In this paper, we present \textbf{\ours}, a system-algorithm co-designed framework for high-resolution, real-time streaming video editing on consumer GPUs, with the following three core designs:
(1) \textbf{Hybrid Diffusion Transformer architecture} introduces softmax attention in part of the blocks to improve local modeling capabilities while preserving the efficiency of linear layers. 
(2) \textbf{Cycle-Reverse Regularization} is a novel training strategy that enforces semantic consistency by predicting source frames from generated content via flow matching, improving temporal consistency without requiring paired long edited videos.
(3) \textbf{Efficient System Co-design} combines fused GDN kernels and Mixed-Precision Quantization (MPQ) optimized for the NVIDIA Blackwell (RTX 5090) architecture. By profiling real-world throughput, our MPQ maximizes Tensor Core utilization while maintaining generation quality.
The resulting system achieves real-time 1280$\times$704 resolution editing at \textbf{24 end-to-end FPS} on a single RTX 5090 GPU, with the DiT core running at \textbf{58 FPS}. Experimental results demonstrate that our co-design approach significantly outperforms existing SOTA methods in both temporal coherence and system throughput.
\end{abstract}

\maketitle

\begin{figure}[H]
  \centering
  \includegraphics[width=\linewidth]{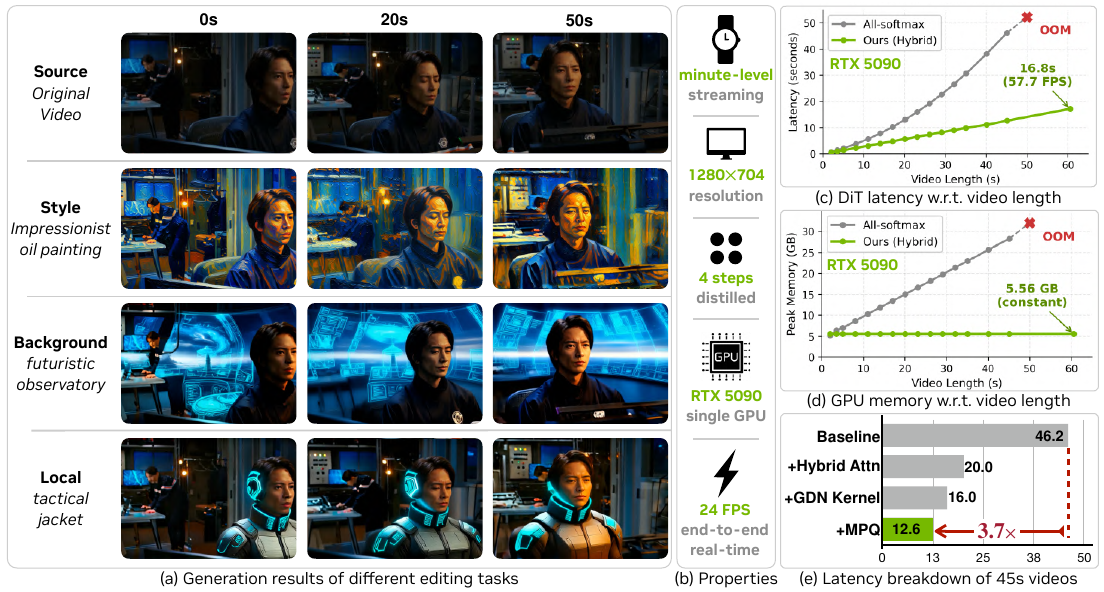}
  \vspace{-5pt}
  \caption{\textbf{Overview of \ours.} Our method supports temporally consistent minute-length video editing at 1280$\times$704 resolution while maintaining bounded memory and real-time throughput on a single RTX 5090.}
  \label{fig:teaser}
\end{figure}

\section{Introduction}
Instruction-guided video editing aims to transform an input video according to a natural-language instruction while preserving the motion and content that should remain unchanged.
Recent video editing systems have made rapid progress on short clips, driven by stronger video diffusion transformers and large-scale synthetic editing data~\cite{jiang2025vace,wu2025insvie,bai2025ditto,he2025openve,liao2025icve,decart2025lucyedit}.
However, real-time streaming V2V editing remains substantially harder.
The model must generate frames in chronological order, maintain long-range appearance consistency across hundreds of frames, and fit within the latency and memory budget of a consumer GPU. These requirements rule out many accurate but bidirectional or full-attention designs that work well for short clips. To this end, we present \ours, a system-algorithm co-designed framework for real-time, high-resolution streaming video editing.

A natural starting point is SANA-Video~\cite{chen2025sanavideo}, the efficient video generation backbone on which our system is built.
SANA-Video replaces vanilla attention with linear attention and introduces block linear attention with a constant-memory state cache, making long high-resolution video generation practical on edge GPUs.
Its long-video extension, LongSANA, further adapts this all-linear backbone to streaming generation through causal chunking, recurrent state reuse, and LongLive-style~\cite{yang2025longlive} streaming long tuning.
This design is attractive for throughput, but it exposes a key limitation for V2V editing: linear attention compresses history into finite recurrent states and is less localized than softmax attention, making it harder to preserve fine-grained source correspondence across chunk boundaries.
In practice, this manifests as visible chunk-to-chunk appearance jumps and temporal flicker in long causal generation (Figure~\ref{fig:la-limitations}).
Conversely, replacing all blocks with softmax attention improves local modeling but becomes memory-prohibitive for long high-resolution streams (Figure~\ref{fig:teaser}(d)).
This tension motivates a hybrid design that allocates expressive softmax attention only where it is most useful, while preserving recurrent linear-state caching in the majority of layers.

\noindent\textbf{Hybrid Diffusion Transformer.} The DiT model of \ours is a hybrid diffusion transformer derived from the SANA-Video's all-linear backbone, where we insert softmax-attention blocks evenly in the all linear backbones and replace the vanilla linear attention with an efficient variant of Gated DeltaNet (GDN)~\cite{yang2024gated}. 
In this hybrid design, \ours leverage GDN blocks to maintain the important global information with dynamic gate and decay. Softmax attention blocks are adopted to efficiently explore the locality and first block consistency with window attention and attention sink. Such design keeps memory constant and generation speed for arbitrary generation length while still keeping long range consistency.
As shown in Figure~\ref{fig:teaser}, compared with all softmax attention variant, the hybrid design only requires 5.56 GB VRAM for long video generation and 3.7 times faster generation speed, enabling real-time execution on the consumer GPUs.

\noindent\textbf{Cycle Consistent Streaming Training.}
Due to the lack of paired long video editing data, we adopt LongLive~\cite{yang2025longlive} as our basic streaming training framework, which uses short video generation teacher to supervise the rollout long video clips with Distribution Matching Distillation (DMD)~\cite{yin2023one,yin2024improved} loss. However, the short teacher supervision lacks the long-range information and therefore may lead drifting when supervise long rollout clips.
In view of this limitation, we introduce Cycle-Reverse Regularization. Starting from the LongLive training strategy, our method adds a reverse editing objective: after generating an edited chunk, the model is required to reconstruct the corresponding source chunk conditioned on a reverse prompt. Since the long video itself is a real-world temporal consistent video, the reverse loss can force the model to learn how to maintain the long range temporal consistency. In addition to the DiT training, we also distill our VAE decoder from bidirectional convolution to causal convolution, supporting streaming decoding.

\noindent\textbf{Efficient System Co-design.}
To enhance the generation efficiency on the consumer GPUs, we explore the inference system design. On the one hand, we improve the efficiency of frame-wise GDN kernel with partition strategy, so that the state matrices can be stored on the GPU's SRAM, leading to \textbf{1.5-2.2x} speed up on different GPU architectures. In addition, since the NVFP4 and FP8 Tensor Core is super efficient on Blackwell architecture while the BF16 precision has the best quality, we design a precision search algorithm inspired from AutoML. By evaluating the trade-off between generation quality and efficiency, we find the best mixed precision strategy for different layers and different blocks, leading to \textbf{59\%} speed up over BF16 on the DiT model with marginal quantization error.

\noindent\textbf{Data Pipeline.} The success of \ours relies on the strong data pipeline for both short videos and long videos. For short videos, we leverage the image editing model to edit the first frame and I2V generation model to obtain the pairs, as well as data verification to filter out low quality pairs. For the long video, since generating consistent pairs is difficult, we use VLM to generate forward and reverse editing prompts for streaming long training and our cycle-reverse regularization.

Our contributions are summarized as follows.
\begin{itemize}
\vspace{-.2cm}
    \item We propose a hybrid diffusion transformer for streaming video editing, combining global information from GDN blocks with strong locality information from softmax-attention blocks for efficient and high quality generation.
    \item We introduce Cycle-Reverse Regularization, which leverages long source videos through a reverse flow-matching objective and improves long-range consistency without requiring paired long edited videos.
    \item We propose a comprehensive editing data generation pipeline for both short video pairs and long video prompts, which is the basic of streaming video editing.
    \item We develop an efficient system design with fused GDN kernels and Mixed-Precision Quantization, achieving 24 end-to-end FPS and 58 DiT FPS on an RTX 5090 GPU.
\end{itemize}

\section{Methodology}

\subsection{Hybrid Architecture}

\begin{figure*}[t]
\centering
\includegraphics[width=\textwidth]{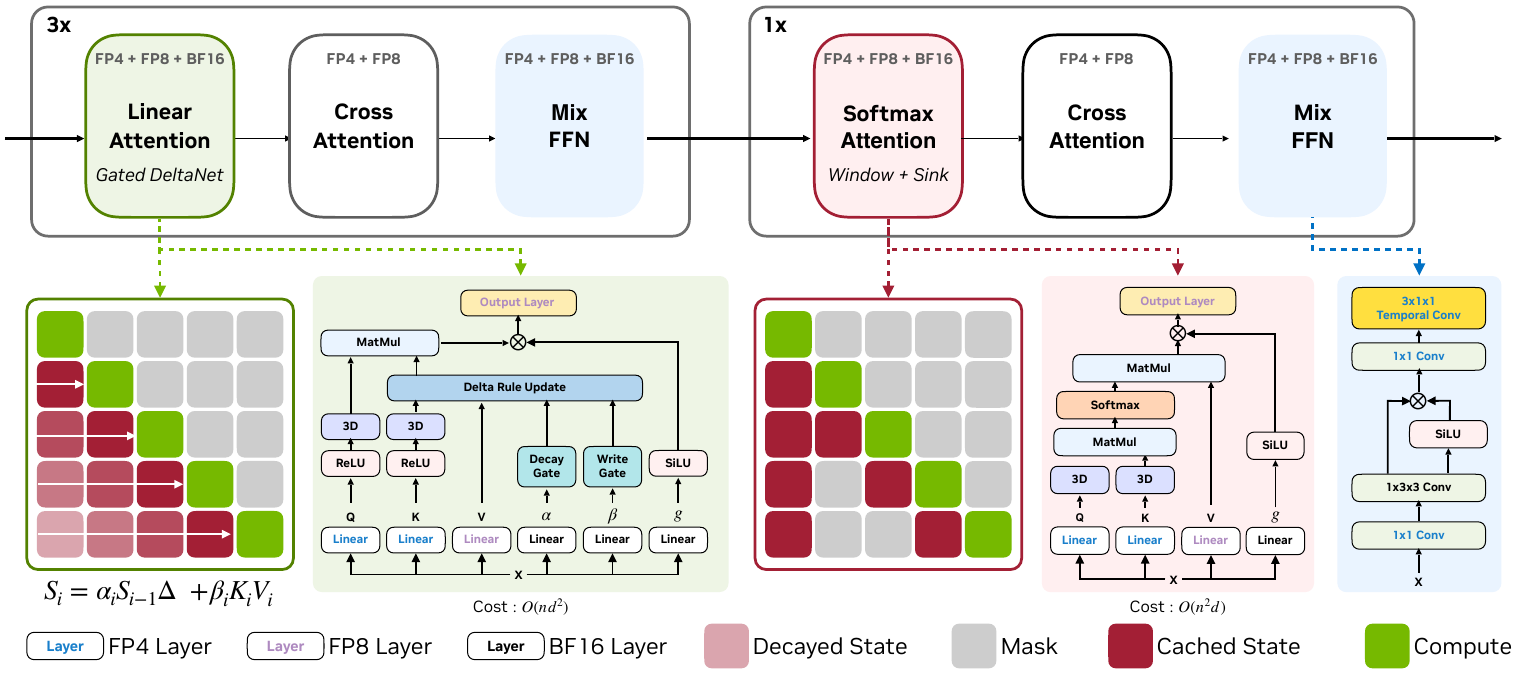}
\vspace{-2em}
\caption{\textbf{Hybrid streaming diffusion transformer.}
\ours interleaves GDN blocks and softmax attention blocks. The two types of blocks focus on different aspect of information with a fixed computational cost. For the detailed block design, each layer will use the most suitable precision for efficiency and generation quality.}
\vspace{-1em}
\label{fig:framework}
\end{figure*}


This section introduces the key architectural design of \ours for minute-length streaming video editing.
Motivated by the complementary limitations of softmax attention and linear attention, we develop a hybrid diffusion transformer enabling long video editing with global consistency and local granularity under fixed GPU memory.

\begin{wrapfigure}{r}{0.33\textwidth}
  \centering
  \includegraphics[width=\linewidth]{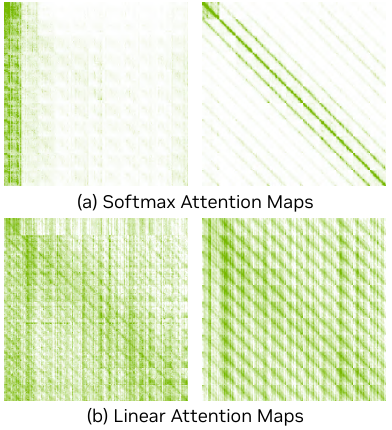}
  \vspace{-2em}
  \caption{Attention map visualization of different heads.}
  \label{fig:attn-map}
\end{wrapfigure}

\noindent\textbf{Limitations of Softmax Attention and Linear Attention.} 
Causal softmax attention provides strong token-level interaction and precise source correspondence, which is crucial for high-quality video-to-video editing.
However, attending to all historical high-resolution video tokens is memory-prohibitive, as the active KV cache grows with video length. 
{In addition, as shown in Figure~\ref{fig:attn-map}(a), softmax attention commonly focuses more on local information. Therefore, restricting softmax attention to a local window with a persistent sink token is applicable while significantly reducing memory cost in long video generation. However, global information is excluded in such attention operation.}
In contrast, linear attention is naturally suitable for streaming because it compresses the generated prefix into a compact recurrent state whose size is independent of the number of chunks.
{Nevertheless, the dense attention maps (Figure~\ref{fig:attn-map}(b)) demonstrate that linear attention does not focus enough on the neighborhood, leading to chunk-boundary flickering (Figure~\ref{fig:la-limitations}).}

We therefore combine the two mechanisms in an interleaved hybrid design as Figure~\ref{fig:framework} shows, following Qwen-Next~\cite{qwen3technicalreport}.
In \ours, most blocks use Gated DeltaNet (GDN) linear attention to maintain compact global memory for consistency, while a small number of blocks use sliding window softmax attention with a persistent sink chunk and a short recent window.
This preserves the efficiency of an all-linear streaming DiT while enabling fine-grained attention to local context.

\noindent\textbf{Frame-wise Gated DeltaNet for Global Accumulated Memory.}
The linear-attention blocks in \ours serve as the global memory pathway for causal streaming.
Instead of token-wise updates or caching raw tokens from all previous chunks, each block maintains a compact recurrent state that is updated frame-by-frame and carried across chunks.
This state summarizes the generated prefix into a fixed-size memory, enabling long-range information propagation under fixed GPU memory.

{Given the latent $x \in \mathbb{R}^{F\times N \times C}$ in each head, we obtain the query $q\in \mathbb{R}^{F\times N \times C}$, key $k \in \mathbb{R}^{F\times N \times C}$, value $v \in \mathbb{R}^{F\times N \times C}$, decay gate $\alpha \in \mathbb{R}^{F}$, write gate $\beta \in \mathbb{R}^{F\times N}$ and output gate $g \in \mathbb{R}^{F\times N \times C}$, where $F$, $N$, and $C$ denote the number of frames, number of tokens in each frame and head dimension respectively. The numerator state of the $f$th frame $S^{kv}_f$ and denominator normalizer state $S^{z}_f$ can be updated with the delta-rule correction~\cite{yang2024gated}:}

\begin{equation}
    S^{kv}_f = \alpha_f S^{kv}_{f-1}(I-\beta_f \hat{k}_f \hat{k}_f^\top) + \beta_f v_f \hat{k}_f^\top, \quad
    S^{z}_f = \alpha_f S^{z}_{f-1}(I-\beta_f k_f k_f^\top) + \beta_f k_f^\top.
    \label{eq:gdn_combined}
\end{equation}

where ${k}_f$ and $\hat{k}_f$ is the key before and after RoPE~\cite{su2021roformer}. 
This update differs from simple accumulation: rather than directly writing $v_f k_f^\top$ into memory, the model only writes the residual.
As a result, the recurrent state behaves as a correction-based memory, which is more robust for long streaming sequences. More details are available in Appendix.~\ref{appendix:gdn}.
The final output is then read from the updated state through a normalized readout followed by an output gate and a linear projection:
\begin{equation}
    o_f
    =
    W_o
    \left(
    g_f \odot
    \frac{
    S^{kv}_f \hat{q}_f
    }{
    (S^z_f)^\top q_f + \epsilon
    }
    \right),
    \label{eq:gdn_out}
\end{equation}
where ${q}_f$ and $\hat{q}_f$ is the query before and after RoPE~\cite{su2021roformer}.
During streaming inference, each Gated DeltaNet block only caches the terminal recurrent states $(S^{kv}, S^z)$ of the previous chunk and uses them to initialize the next chunk.
Therefore, the memory cost of the linear-attention pathway remains independent of the number of streamed chunks.
In our hybrid architecture, these Gated DeltaNet blocks provide a compact global accumulated memory.

\noindent\textbf{Softmax blocks for local window and sink refinement.}
GDN blocks provide efficient global memory and softmax attention blocks help for more locality information. Instead of using all the frames for attention, 
each chunk attends to a constrained context consisting of itself, a persistent sink, and a local window.
The sink chunk provides a stable visual anchor shared across the stream, while the recent window preserves high-resolution local evidence from neighboring chunks.
This design restores precise local matching with fixed GPU memory.

\subsection{Long-term Consistency Training}
\begin{figure*}[t]
\centering
\includegraphics[width=\textwidth]{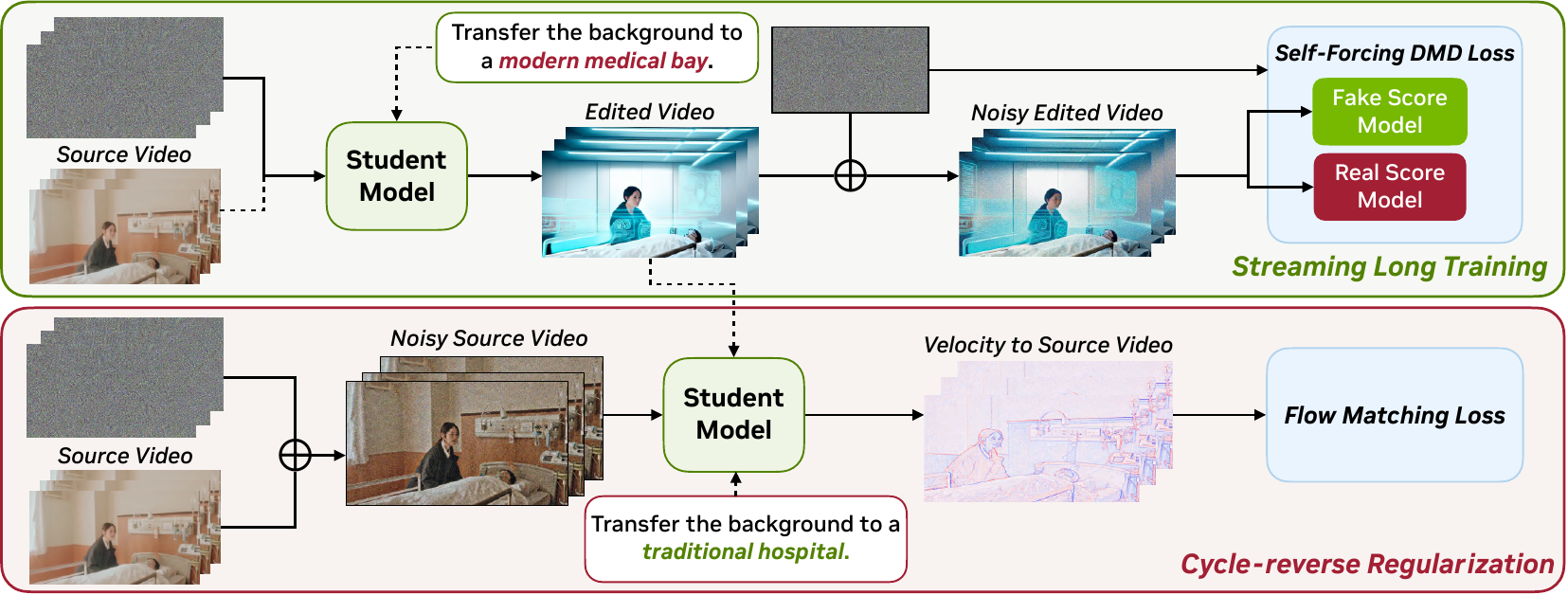}
\caption{\textbf{Streaming Long Training and Cycle-Reverse Regularization.}
The forward streaming branch uses streaming long training with DMD loss.
The reverse branch reuses the generated edited chunk as visual condition and applies a reverse instruction to reconstruct the original source chunk with a flow-matching loss, which improves temporal consistency without paired long edited targets.}
\vspace{-1em}
\label{fig:cycle_reverse}
\end{figure*}

\noindent\textbf{Streaming long training.}
We train \ours under the same causal streaming protocol used during inference.
Given a long source video and an editing prompt, the model generates edited latent chunks autoregressively.
Upon generating each chunk, the streaming cache is updated with the generated chunk, and a LongLive-style distribution matching distillation (DMD) self-forcing loss~\cite{yang2025longlive,huang2025selfforcing} is applied to the current chunk.
mitigates the discrepancy between short-clip training and minute-length streaming inference, enabling the model to learn from its own generated history.

\noindent\textbf{Cycle-reverse regularization.}
While streaming long training improves long-horizon generation, video-to-video editing further requires the edited stream to preserve the source structure and motion over time.
Although minute-length source videos are relatively easy to collect, paired target videos that faithfully follow the editing instruction are rarely available.
We therefore introduce a cycle-reverse regularization strategy that uses only long source videos without paired edited videos.

Specifically, the model first performs a forward streaming edit from the source video to the target domain according to the editing prompt.
Then, the generated edited chunk is used as the visual condition for a reverse edit, guided by a reverse prompt that describes how to recover the original source domain, as shown in Figure~\ref{fig:cycle_reverse}.
The reverse pass is trained with a flow-matching objective toward the original source chunk.
Overall, streaming long training teaches the model to follow editing instructions under causal rollout and cycle-reverse regularization encourages the edited video to preserve long-term structure and motion consistency learned from real-world source videos.

\begin{wrapfigure}{r}{0.4\textwidth}
  \centering
  \vspace{-1em}
  \includegraphics[width=\linewidth]{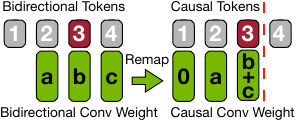}
  \vspace{-1.5em}
  \caption{Remapping weights in VAE.}
  \label{fig:causal-conv-in-vae}
  \vspace{-1em}
\end{wrapfigure}

\noindent\textbf{Causal VAE Distillation.}
The video VAE decoder is another bottleneck for streaming video editing. The original LTX2 VAE decoder uses bidirectional temporal context, while streaming inference cannot access future latent frames.
We convert it into a causal decoder by replacing symmetric temporal padding with left-only padding, so each output frame only depends on current and previous latent frames.
For stable initialization, we reuse the pretrained LTX2 VAE weights and remap each temporal $3$-tap convolution from bidirectional weights $[a,b,c]$ to causal weights $[0,a,b+c]$, preserving the previous-frame and current-frame roles while folding the unavailable future-frame contribution into the current frame, as shown in Figure~\ref{fig:causal-conv-in-vae}.
We then distill the causal decoder from the original bidirectional decoder.
The training objective combines Charbonnier reconstruction loss~\cite{charbon1994loss}, perceptual loss~\cite{johnson2016perceptualloss}, Haar wavelet high-frequency loss~\cite{ye2025ultraflux, zhang2025diffusion4k}, and intermediate decoder feature distillation~\cite{zou2025turbovaed}:
\begin{equation}
    \mathcal{L}_{\mathrm{vae}}
    =
    \lambda_{\mathrm{charb}}\mathcal{L}_{\mathrm{charb}}
    + \lambda_{\mathrm{perc}}\mathcal{L}_{\mathrm{perc}}
    + \lambda_{\mathrm{haar}}\mathcal{L}_{\mathrm{haar}}
    + \lambda_{\mathrm{distill}}\mathcal{L}_{\mathrm{distill}}.
    \label{eq:causal-vae-objective}
\end{equation}

\subsection{Efficient System Co-design}
\begin{figure*}[t]
    \centering
    \begin{subfigure}[b]{0.48\textwidth}
        \centering
        \includegraphics[width=\textwidth]{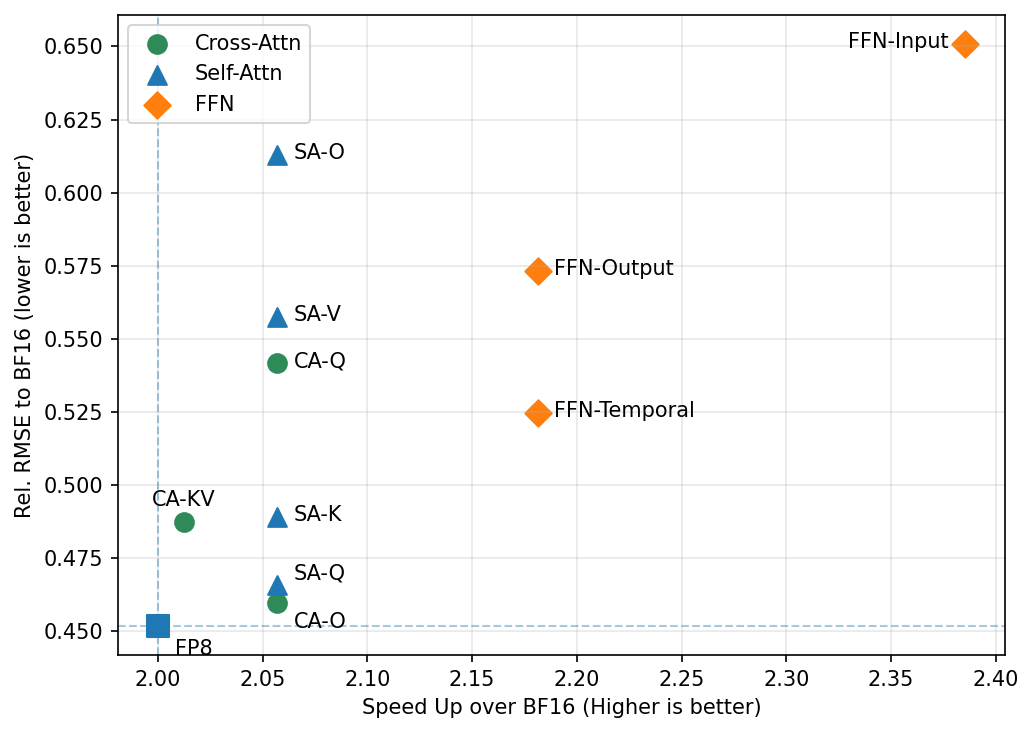} 
        \caption{Per-layer quantization policy search}
        \label{fig:mpq_relrmse_layer}
    \end{subfigure}
    \hfill 
    \begin{subfigure}[b]{0.48\textwidth}
        \centering
        \includegraphics[width=\textwidth]{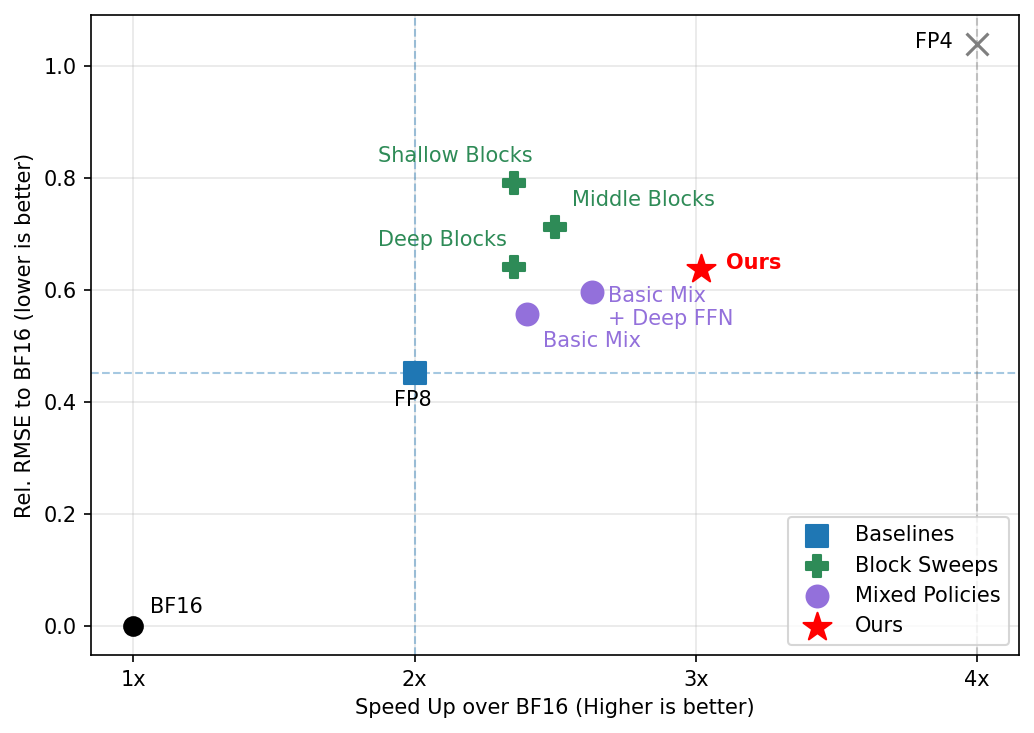} 
        \caption{Per-block and mixed quantization policy search}
        \label{fig:mpq_relrmse_block}
    \end{subfigure}
    \vspace{-.05in}
    \caption{\textbf{Mixed-Precision Quantization Policy Search on Relative RMSE}. For each subfigure, the x-axis is the estimated speed up and the y-axis is the quantization error over the BF16 reference. The selected mixed-precision policy achieves the best trade-off between efficiency and quality.}
    \label{fig:mpq_relrmse}
    \vspace{-.1in}
\end{figure*}

\noindent\textbf{Efficient GDN Kernel.}
The recurrent GDN update is algorithmically efficient, but a literal
implementation is not hardware efficient. For both H100 and RTX 5090, the
recurrent state $(S_f,z_f)$ is small enough to keep in on-chip storage, while
the per-frame $Q_f,K_f,V_f$ activations span hundreds of spatial tokens and
must be streamed from HBM. Following the IO-aware principle of
FlashAttention~\cite{dao2022flashattention, dao2023flashattention2}, we tile the spatial dimension so that each block of
activations is loaded once, reduced into compact frame-level summaries, and
then discarded.

We rewrite the update by separating the spatial reduction from the temporal
recurrence:
\begin{align}
    P_f &= I - K_f^{\top}\operatorname{diag}(\beta_f)K_f, &
    A_f &= K_f^{\top}\operatorname{diag}(\beta_f)V_f, \\
    S_f &= \alpha_f P_f S_{f-1} + A_f .
\end{align}
The vector path computes
$P^z_f = I-K_f^\top\operatorname{diag}(\beta_f)K_f$ and
$b_f=K_f^\top\beta_f$, then
$z_f=\alpha_f P^z_f z_{f-1}+b_f$. This decomposition exposes most of the work
as frame-parallel: $P_f,A_f,P^z_f,b_f$ are computed independently per frame and head, while only the compact scan over $(S_f,z_f)$ remains sequential.

Our Triton implementation uses a three-phase chunkwise pipeline. Phase A
computes the frame summaries with blocked reductions over the spatial
dimension. Phase B performs the short recurrent scan over frames while keeping
the state in on-chip storage. Phase C streams $Q_f$ blocks and applies the
compact states to produce the output. For bidirectional GDN, we exploit the
linearity of Phase C and combine the forward and reverse histories before the
output pass, avoiding a second output kernel and a second read of $Q_f$. This
chunkwise design consistently gives large layer-level speedups and roughly $1.5\times$--$2.2\times$
end-to-end sampling speedups over the PyTorch reference.

\noindent\textbf{Mixed-Precision Quantization Policy Search.}
Quantization is further explored in this paper to support the efficiency on consumer GPUs.
NVFP4 quantization is fast on Blackwell architecture but not uniformly reliable, for two main reasons: (1) some layers are more sensitive and should remain in higher precision, such as the patch embedding layer and output layer; (2) some layers has very few parameters with small computational cost, so the quantization overhead exceeds the kernel speedup, such as gate layer.  
We therefore search at layer-type and block-position granularity instead of assigning the same precision to the whole DiT.
We first set the most sensitive or unsupported layers to BF16, including the patch embedding layer, output layer, timestep embedding layer, the gates in attention blocks and depth-wise convolution in MixFFN. For the remaining layers, we start from an FP8 base and apply group wise NVFP4 on it.
The searchable FP4 groups include per-layer groups: self-attention $Q/K/V/O$, cross-attention $Q/KV/O$, FFN input and output projections, and temporal FFN projections; block-wise groups: shallow (0-5), middle (6-13) and deep blocks (14-19).
For each candidate $\pi$, we run a 30-second streaming video generation with the same noise on a calibration set.
Let $\hat{x}^{\mathrm{ref}}_{0}$ be the BF16 output latent and $\hat{x}^{\pi}_{0}$ be the output latent under policy $\pi$.
We score the candidate by latent relative RMSE and LPIPS~\cite{johnson2016perceptualloss}:
\begin{equation}
    \mathrm{RelRMSE}(\pi)
    =
    \frac{
    \sqrt{\frac{1}{N}\left\|
    \hat{x}^{\pi}_{0}
    -
    \hat{x}^{\mathrm{ref}}_{0}
    \right\|_2^2}
    }{
    \mathrm{Std}\!\left(\hat{x}^{\mathrm{ref}}_{0}\right) + \epsilon
    }.
\end{equation}
Figure~\ref{fig:mpq_relrmse} and Figure~\ref{fig:mpq_lpips} illustrate how each group influences efficiency and quality. We make the following observations. 
\textbf{First}, the key and query of self-attention (SA-K, SA-Q) and the output projection of cross-attention (CA-O) sit nearly on top of the FP8 baseline while still delivering a ${\sim}2.06\times$ speedup, so they are the safest layers to demote and form our \emph{Basic Mix}, joined by FFN-Temporal -- the only FFN component whose quantization error is negligible relative to its efficiency gain. 
\textbf{Second}, the remaining attention projections (SA-V, SA-O, CA-Q) move noticeably above the FP8 horizontal, indicating that the value and post-mixing pathway is sensitive to FP4 noise; we keep them in FP8. 
\textbf{Third}, CA-KV sits closest to FP8 on both axes but covers only $1\%$ of the DiT FLOPs, so demoting it brings little speedup and we exclude it from the policy. 
\textbf{Fourth}, the FFN input/output point-wise convolutions are the heaviest modules in the network ($\sim$49\% of FLOPs combined); demoting them globally collapses quality.
\textbf{However}, the block-range sweep reveals a clear depth ordering: shallow blocks are the most fragile, while deep and middle blocks tolerate FP4 well,  opening a depth-restricted demotion path. 
To this end, the resulting policy (red star in Figure~\ref{fig:mpq_relrmse}) uses FP4 for the most robust groups, including cross-attention output, self-attention query and key, temporal FFN projection in all blocks as well as FFN input and output projections in middle and deep blocks. The unsupported and most sensitive layers listed at the start of this section (patch embedding, output projection, timestep embedding, attention gates, MixFFN depth-wise convolution) remain in BF16, and all other linear layers not assigned to FP4 stay in FP8.
The mixed-precision policy gains 3x efficiency on the quantized layers and 1.59x efficiency on DiT latency over BF16 baseline. The detailed analysis of each layer and block are available in Appendix~\ref{appendix:mpq}.

\section{Data Pipeline} 
\begin{figure*}[t]
\centering
\includegraphics[width=\textwidth]{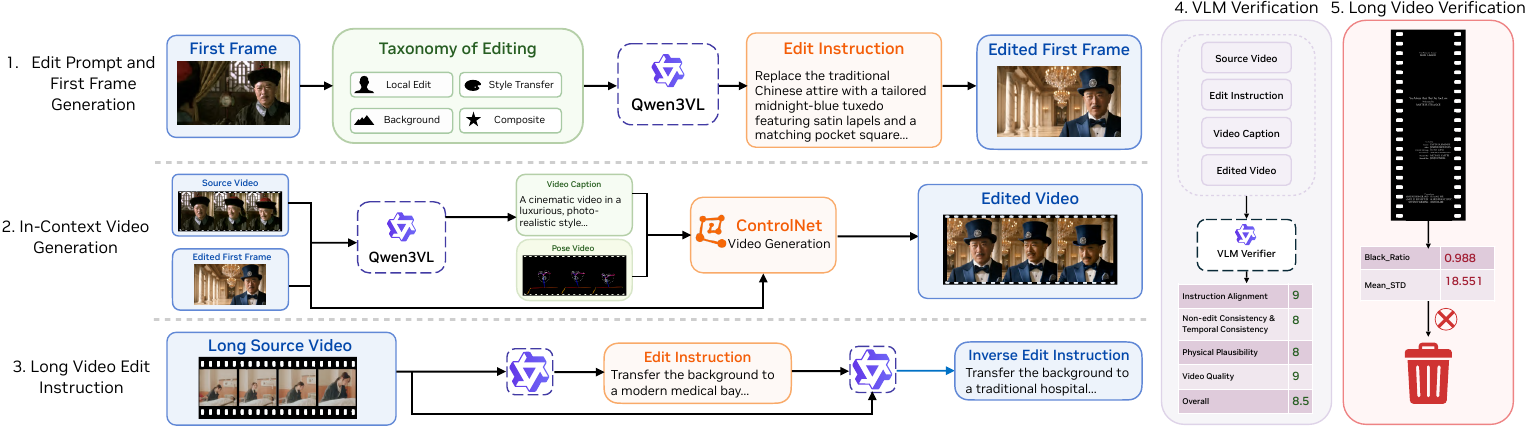}
\caption{\textbf{Data Pipeline.}
For short videos, we sample taxonomy-guided edit instructions, edit the first frame as a visual anchor, and generate edited videos by conditioning on the source video, edited first frame, caption, and pose video.
For long videos, we generate paired forward and inverse instructions from source videos to support streaming long training and cycle-reverse regularization.
Both short and long samples are filtered by VLM-based verification, and long videos are additionally screened for invalid black-frame segments.}
\vspace{-.1in}
\label{fig:data_pipeline}
\end{figure*}

Since one of the main downstream tasks of \ours is live broadcasting, we introduce our data pipeline to further generate high-quality human-centric video editing pairs as well as the editing prompts for long videos.
Figure~\ref{fig:data_pipeline} summarizes the complete pipeline, including short-video pair construction, long-video instruction construction, and VLM-based filtering. The details of our data generation pipeline are available at Appendix~\ref{appendix:data}.

\section{Experiments}

\subsection{Implementation Details}
\label{sec:implementation}
\ours is a 2B hybrid diffusion transformer with LTX2 VAE~\cite{hacohen2026ltx2} (compression ratio 32$\times$32$\times$8). The DiT contains 20 transformer blocks with evenly inserted 5 softmax attention blocks and 15 GDN blocks. The condition video latent is concatenated to the latent on the channel dimension. The model is trained by two stages on 1280$\times$704 resolution: bidirectional short training and streaming long training. 
In the bidirectional short training stage, we use bidirectional GDN with forward and backward scan. In the streaming long training stage, 3 latent frames are regarded as a chunk. The forward scan with previous cache and backward scan within the chunk is adopted by GDN. The streaming model is distilled to 4 steps.
With the algorithm and system co-design, \ours achieves 24 end-to-end FPS and 58 DiT FPS on an RTX 5090 GPU.
For quantitative evaluation, we use the five OpenVE-Bench~\cite{he2025openve} pixel-aligned categories: global style, background change, local change, local remove, and local add.

\subsection{Performance Comparison and Analysis}

\begin{table*}[t]
\centering
\caption{OpenVE-Bench comparison on the five spatially aligned edit categories used in our evaluation. The \rankfirst{color} and \ranksecond{color} highlights indicate the best and second-best quality scores, respectively. Latency is measured with batch size 1, while throughput is measured with batch inference on eight H100 GPUs for 81-frame 1280$\times$704 videos. $\dagger$ indicates the step distilled model.}
\label{tab:performance_comparison}
\resizebox{\linewidth}{!}{
\begin{tabular}{lrrrrrrrrrr}
\toprule
\multirow{2}{*}{Methods} & \multirow{2}{*}{\#Params.} & \multirow{2}{*}{\#Reso.} & Latency & Throughput & \multirow{2}{*}{~~Avg.} & Global & BG & Local & Local & ~Local \\
 & & & (s) & (FPS) & & Style & Change & Change & Remove & Add \\
\midrule
VACE~\cite{jiang2025vace}          & 14B  & 1280x720 & 1991 & 0.3 & 1.57 & 1.49 & 1.55 & 2.07 & 1.46 & 1.26 \\
OmniVideo~\cite{tan2025omni}     & 1.3B & 640x352  & ---   & --- & 1.19 & 1.11 & 1.18 & 1.14 & 1.14 & 1.36 \\
InsViE~\cite{wu2025insvie}        & 2B   & 720x480  & 750   & 0.9 & 1.45 & 2.20  & 1.06 & 1.48 & 1.36 & 1.17 \\
Lucy-Edit~\cite{decart2025lucyedit}     & 5B   & 1280x704 & {97}    & {6.7} & 2.22 & 2.27 & 1.57 & \rankfirst{3.20}  & 1.75 & \rankfirst{2.30}  \\
ICVE~\cite{liao2025icve}          & 13B  & 384x240  & 6051   & 0.2 & 2.18 & 2.22 & 1.62 & 2.57 & \rankfirst{2.51} & {1.97} \\
DITTO~\cite{bai2025ditto}          & 14B  & 832x480  & 1971  & 0.3 & 2.13 & \rankfirst{4.01} & 1.68 & 2.03 & 1.53 & 1.41 \\
OpenVE-Edit~\cite{he2025openve}        & 5B   & 1280x704 & {97}    & {6.7} & \ranksecond{2.50} & 3.16 & \rankfirst{2.36} & {2.98} & 1.85 & \ranksecond{2.15} \\
\midrule
\textbf{\ours} & 2B & 1280x704 & \ranksecond{20} & \ranksecond{32.4} & \rankfirst{2.62} & {3.48} & \ranksecond{2.29} & \rankfirst{3.20} & \ranksecond{2.27} & 1.88 \\
\textbf{\ours $\dagger$} & 2B & 1280x704 & \rankfirst{1} & \rankfirst{762.8} & {2.42} & \ranksecond{3.60} & {1.82} & \ranksecond{3.10} & {1.78} & 1.82 \\
\bottomrule
\end{tabular}
}
\end{table*}

\begin{figure*}[t]
\centering
\includegraphics[width=\textwidth]{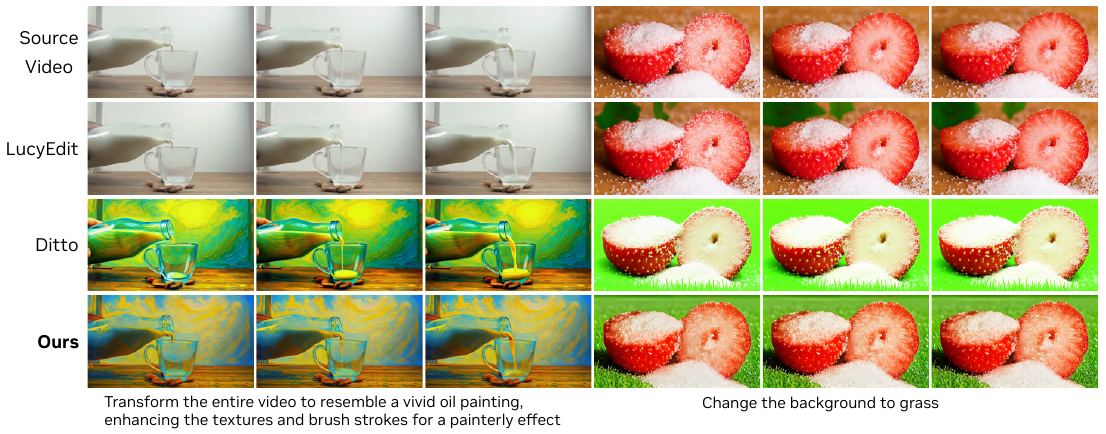}
\caption{\textbf{Qualitative comparison with prior video editors.}
\ours preserves source motion and non-edited content while following the edit instructions.}
\label{fig:sota_qualitative}
\vspace{-1em}
\end{figure*}

Table~\ref{tab:performance_comparison} compares \ours with representative instruction-guided video editing systems on the five OpenVE-Bench categories used in our evaluation.
\ours first stage model is a bidirectional model on short videos, and it achieves the state-of-the-art performance on this benchmark, with nearly 2.5x smaller model size and 5x faster throughput than the previous best method (OpenVE~\cite{he2025openve}).
The streaming and distilled version of our model targets deployment on consumer GPUs. However, even with distillation, \ours can still achieve comparable performance with previous methods and has more than 100x higher throughput.
Qualitative comparisons are shown in Figure~\ref{fig:sota_qualitative} and more qualitative long video results of \ours are available in Figure~\ref{fig:more-results}.

\subsection{Ablation Studies}

\begin{table}[ht]
  \centering
  \begin{minipage}[t]{0.53\linewidth}
    \centering
    \caption{Latency (s) analysis of the efficient system co-design on an RTX 5090 for 1-minute video.}
    \label{tab:system_runtime_ablation}
    \resizebox{\linewidth}{!}{
      \begin{tabular}{lccccc}
      \toprule
      System  & VAE  & VAE & {DiT} & End-to-end & DiT \\
      Setting & Encode & Decode & Latency & Latency & Speedup \\
      \midrule
      BF16 & \multirow{3}{*}{15.4} & \multirow{3}{*}{8.5}  & 26.8 & 50.7 & 1.00$\times$ \\
       + GDN Kernel & &  & 21.9 & 45.8 & 1.22$\times$ \\
       + MPQ & & & 16.8 & 40.7 & 1.59$\times$ \\
      \bottomrule
      \end{tabular}
    }
  \end{minipage}
  \hfill 
  \begin{minipage}[t]{0.46\linewidth}
    \centering
    \caption{Causal VAE Distillation Comparison.}
    \vspace{1.1em}
    \label{tab:causal_vae}
    \resizebox{\linewidth}{!}{
      \begin{tabular}{lccc}
      \toprule
      Methods & PSNR & LPIPS & SSIM\\
      \midrule
      Bidirectional & 32.98 & 0.0274 & 0.923 \\
      Causal (Before Training) & 24.66 & 0.132 & 0.785 \\
      \textbf{Causal (After Training)} & 32.14 & 0.0336 & 0.911 \\
      \bottomrule
      \end{tabular}
    }
  \end{minipage}
\end{table}

\noindent\textbf{Efficient System Co-design.} Efficient system co-design is crucial to achieve real-time end-to-end generation on consumer GPUs. In Table~\ref{tab:system_runtime_ablation}, we break down the deployment optimizations over the BF16 baseline. Specifically, the lossless efficient GDN kernel improves the DiT latency by 22\%, and Mixed-Precision Quantization (MPQ) further reduces the DiT latency to 16.8s, yielding a 1.59$\times$ DiT speedup over the BF16 baseline and achieving 24 FPS for end-to-end generation.
These results demonstrate the importance and effectiveness of our algorithm-system co-design.

\begin{figure*}[t]
\centering
\includegraphics[width=0.98\textwidth]{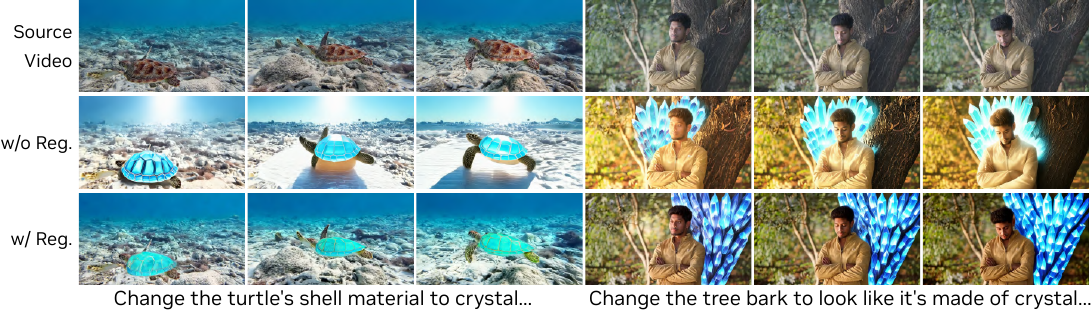}
\caption{\textbf{Effect of Cycle-Reverse Regularization.}
Adding cycle-reverse regularization improves the preservation of source motion and non-edited regions while keeping the video temporally stable.}
\label{fig:cycle_reverse_ablation}
\end{figure*}

\begin{figure*}[t]
\centering
\vspace{-.1in}
\includegraphics[width=\textwidth]{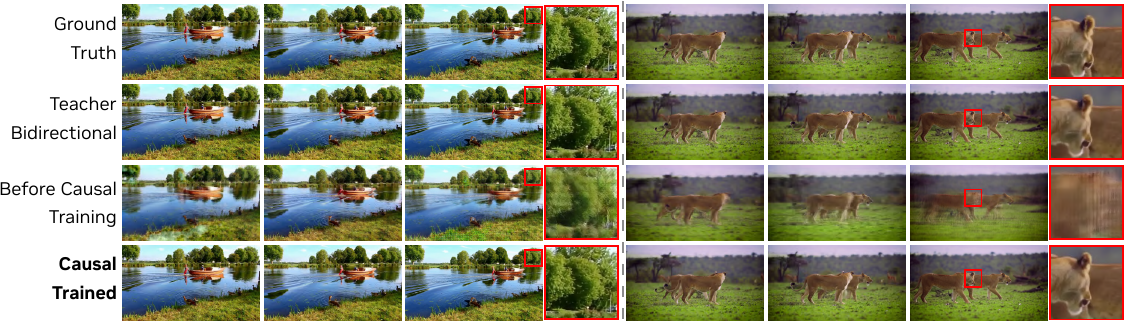}
\caption{\textbf{Causal VAE distillation.}
Our decoder is comparable with the bidirectional teacher model.}
\label{fig:causal_vae}
\vspace{-0.5em}
\end{figure*}

\noindent\textbf{Cycle-Reverse Regularization.} 
Cycle-Reverse Regularization aims to improve temporal consistency when the paired long videos are not available. As shown in the first case of Figure~\ref{fig:cycle_reverse_ablation}, cycle-reverse regularization can keep the non-edited region consistent with the source video. In the second case, the edited ``crystal tree'' changes in the later frames when regularization is not applied but our model keeps it consistent throughout the whole long video.

\noindent\textbf{Causal VAE.}
We compare the fine-tuned causal VAE decoder against the original decoder with causal and non-causal decoding in Figure~\ref{fig:causal_vae}. 
Direct causal conversion introduces visible blur and loss of fine detail because the decoder no longer has access to future latent frames.
After causal decoder distillation, the streaming VAE recovers sharper textures and object boundaries, achieving comparable performance with the bidirectional teacher (Figure~\ref{fig:causal_vae} and Table~\ref{tab:causal_vae}). 

\section{Related Work}
\label{appendix:related-work}

\noindent\textbf{Hybrid Linear-Softmax Architectures.}
Diffusion Transformers (DiTs) have become the dominant paradigm for high-quality image~\cite{xie2024sana,xie2025sana,chen2025sanasprint,ye2025ultraflux,wu2025qwenimagetechnicalreport} and video generation~\cite{wan2025wan,gao2025seedance,hacohen2026ltx2}. 
However, full attention incurs quadratic complexity, leading to prohibitive memory and latency costs for long sequences.
To improve efficiency, prior work such as SANA-Video~\cite{chen2025sanavideo,zhu2026sana} replaces full attention with linear attention and SLA~\cite{sla} replaces softmax attention with sparse attention and linear attention. 
While effective for scaling, pure linear attention often degrades fine-grained modeling, especially for short-range dependencies.
Recent LLM studies suggest that neither full nor linear attention alone achieves the best efficiency–quality tradeoff. 
Hybrid designs therefore interleave linear and softmax attention to combine scalable context modeling with periodic refinement, as seen in Kimi Linear~\cite{kimiteam2025kimilinear} and Qwen3-Next/3.5~\cite{qwen3technicalreport,qwen2026qwen35}.
Inspired by the success of hybrid architectures in LLMs, we introduce a hybrid diffusion transformer that combines the local modeling strength of softmax attention with the global recurrent memory of linear attention for efficient, high-quality video generation.

\noindent\textbf{Video Editing.}
Instruction-guided video editing has recently shifted from tuning-free attention manipulation and mask-based control toward large supervised video editing models.
VACE~\cite{jiang2025vace} unifies video creation and editing through a video condition interface and context adapter.
InsViE~\cite{wu2025insvie} and Ditto~\cite{bai2025ditto} emphasize scalable synthetic paired-data construction for instruction-based editing.
OpenVE~\cite{he2025openve} further introduces OpenVE-3M and OpenVE-Bench, a unified benchmark with diverse spatially aligned and non-spatially aligned editing categories.
Other recent systems, including OmniVideo~\cite{tan2025omni}, ICVE~\cite{liao2025icve}, and Lucy-Edit~\cite{decart2025lucyedit}, improve instruction following and local edit fidelity through stronger multimodal conditioning, in-context learning, or open-weight video editing backbones.
These methods mainly target offline short-clip editing.
In contrast, \ours focuses on real-time streaming V2V editing and achieves real-time video editing on a consumer GPU.

\noindent\textbf{Long Video Generation.}
With the development of stronger video generation models, long video generation has attracted increasing attention.
Diffusion Forcing~\cite{chen2024diffusionforcing}, Self-Forcing~\cite{huang2025selfforcing}, and Causal-Forcing~\cite{zhu2026causalforcing} explore causal and autoregressive generation paradigms for producing long videos beyond the context length of pre-trained short-video generation models.
Furthermore, LoL~\cite{cui2026lollongerlongerscaling} studies the misbehavior of RoPE in long-video generation and proposes scaling strategies for longer temporal contexts. 
LongLive~\cite{yang2025longlive,chen2026longlive20nvfp4parallelinfrastructure} shows that causal frame-level generation, KV recaching, streaming long tuning, and short-window attention with a frame sink can enable real-time interactive long video generation.
\ours uses LongLive as the base streaming training framework and proposes cycle-reverse regularization to further improve temporal consistency.

\section{Conclusion}
\label{sec:conclusion}
We present \ours, a system–algorithm co-designed framework for real-time, high-resolution streaming video-to-video editing on consumer GPUs. By combining a hybrid diffusion transformer with Cycle-Reverse Regularization and an efficient system design, our approach achieves a favorable balance between temporal consistency, visual quality, and inference efficiency under streaming constraints. Empirical results demonstrate that SANA-Streaming enables minute-length video editing at real-time speed while maintaining competitive editing quality. This work provides a practical step toward interactive video editing systems and highlights the importance of jointly addressing modeling, training, and system challenges in long-form generative tasks.

\noindent\textbf{Limitations.}
While Cycle-Reverse Regularization mitigates the scarcity of paired data, the shortage of high-quality long video editing samples still hinders temporal consistency in complex scenarios.
In addition, like other instruction-guided generative models, SANA-Streaming may produce inconsistent or incorrect edits under ambiguous or underspecified instructions. The model lacks explicit mechanisms to resolve ambiguity or guarantee faithful execution of user intent, which can lead to unpredictable outputs in challenging cases.

\clearpage
\bibliographystyle{plain} 
\bibliography{reference}

@misc{chen2026longlive20nvfp4parallelinfrastructure,
      title={LongLive-2.0: An NVFP4 Parallel Infrastructure for Long Video Generation},
      author={Yukang Chen and Luozhou Wang and Wei Huang and Shuai Yang and Bohan Zhang and Yicheng Xiao and Ruihang Chu and Weian Mao and Qixin Hu and Shaoteng Liu and Yuyang Zhao and Huizi Mao and Ying-Cong Chen and Enze Xie and Xiaojuan Qi and Song Han},
      year={2026},
      eprint={2605.18739},
      archivePrefix={arXiv},
      primaryClass={cs.CV},
      url={https://arxiv.org/abs/2605.18739},
}

@article{zhu2026sana,
  title={SANA-WM: Efficient Minute-Scale World Modeling with Hybrid Linear Diffusion Transformer},
  author={Zhu, Haoyi and Liu, Haozhe and Zhao, Yuyang and Ye, Tian and Chen, Junsong and Yu, Jincheng and He, Tong and Han, Song and Xie, Enze},
  journal={arXiv preprint arXiv:2605.15178},
  year={2026}
}

@misc{aigc_apps_VideoX_Fun_2026,
  author = {aigc-apps},
  title = {VideoX-Fun: A Video Generation Pipeline for Diffusion Transformer},
  year = {2026},
  publisher = {GitHub},
  url = {https://github.com/aigc-apps/VideoX-Fun}
}

@article{inan2023llama,
  title={Llama guard: Llm-based input-output safeguard for human-ai conversations},
  author={Inan, Hakan and Upasani, Kartikeya and Chi, Jianfeng and Rungta, Rashi and Iyer, Krithika and Mao, Yuning and Tontchev, Michael and Hu, Qing and Fuller, Brian and Testuggine, Davide and others},
  journal={arXiv preprint arXiv:2312.06674},
  year={2023}
}

@article{xie2025sana,
  title={Sana 1.5: Efficient scaling of training-time and inference-time compute in linear diffusion transformer},
  author={Xie, Enze and Chen, Junsong and Zhao, Yuyang and Yu, Jincheng and Zhu, Ligeng and Wu, Chengyue and Lin, Yujun and Zhang, Zhekai and Li, Muyang and Chen, Junyu and others},
  journal={arXiv preprint arXiv:2501.18427},
  year={2025}
}

@inproceedings{chen2025sanasprint,
  title={Sana-sprint: One-step diffusion with continuous-time consistency distillation},
  author={Chen, Junsong and Xue, Shuchen and Zhao, Yuyang and Yu, Jincheng and Paul, Sayak and Chen, Junyu and Cai, Han and Han, Song and Xie, Enze},
  booktitle={Proceedings of the IEEE/CVF International Conference on Computer Vision},
  pages={16185--16195},
  year={2025}
}

@inproceedings{yin2023one,
  title={One-step diffusion with distribution matching distillation},
  author={Yin, Tianwei and Gharbi, Micha{\"e}l and Zhang, Richard and Shechtman, Eli and Durand, Fredo and Freeman, William T and Park, Taesung},
  booktitle={Proceedings of the IEEE/CVF conference on computer vision and pattern recognition},
  pages={6613--6623},
  year={2024}
}

@article{yin2024improved,
  title={Improved distribution matching distillation for fast image synthesis},
  author={Yin, Tianwei and Gharbi, Micha{\"e}l and Park, Taesung and Zhang, Richard and Shechtman, Eli and Durand, Fredo and Freeman, William T},
  journal={Advances in neural information processing systems},
  volume={37},
  pages={47455--47487},
  year={2024}
}

@article{qwen3technicalreport,
  title={Qwen3 technical report},
  author={Yang, An and Li, Anfeng and Yang, Baosong and Zhang, Beichen and Hui, Binyuan and Zheng, Bo and Yu, Bowen and Gao, Chang and Huang, Chengen and Lv, Chenxu and others},
  journal={arXiv preprint arXiv:2505.09388},
  year={2025}
}

@article{chen2025sanavideo,
  title={Sana-video: Efficient video generation with block linear diffusion transformer},
  author={Chen, Junsong and Zhao, Yuyang and Yu, Jincheng and Chu, Ruihang and Chen, Junyu and Yang, Shuai and Wang, Xianbang and Pan, Yicheng and Zhou, Daquan and Ling, Huan and others},
  journal={arXiv preprint arXiv:2509.24695},
  year={2025}
}

@article{yang2025longlive,
  title={Longlive: Real-time interactive long video generation},
  author={Yang, Shuai and Huang, Wei and Chu, Ruihang and Xiao, Yicheng and Zhao, Yuyang and Wang, Xianbang and Li, Muyang and Xie, Enze and Chen, Yingcong and Lu, Yao and others},
  journal={arXiv preprint arXiv:2509.22622},
  year={2025}
}

@inproceedings{jiang2025vace,
  title={Vace: All-in-one video creation and editing},
  author={Jiang, Zeyinzi and Han, Zhen and Mao, Chaojie and Zhang, Jingfeng and Pan, Yulin and Liu, Yu},
  booktitle={Proceedings of the IEEE/CVF International Conference on Computer Vision},
  pages={17191--17202},
  year={2025}
}

@inproceedings{wu2025insvie,
  title={Insvie-1m: Effective instruction-based video editing with elaborate dataset construction},
  author={Wu, Yuhui and Chen, Liyi and Li, Ruibin and Wang, Shihao and Xie, Chenxi and Zhang, Lei},
  booktitle={Proceedings of the IEEE/CVF International Conference on Computer Vision},
  pages={16692--16701},
  year={2025}
}

@article{bai2025ditto,
  title={Scaling instruction-based video editing with a high-quality synthetic dataset},
  author={Bai, Qingyan and Wang, Qiuyu and Ouyang, Hao and Yu, Yue and Wang, Hanlin and Wang, Wen and Cheng, Ka Leong and Ma, Shuailei and Zeng, Yanhong and Liu, Zichen and others},
  journal={arXiv preprint arXiv:2510.15742},
  year={2025}
}

@article{he2025openve,
  title={OpenVE-3M: A Large-Scale High-Quality Dataset for Instruction-Guided Video Editing},
  author={He, Haoyang and Wang, Jie and Zhang, Jiangning and Xue, Zhucun and Bu, Xingyuan and Yang, Qiangpeng and Wen, Shilei and Xie, Lei},
  journal={arXiv preprint arXiv:2512.07826},
  year={2025}
}

@article{tan2025omni,
  title={Omni-video: Democratizing unified video understanding and generation},
  author={Tan, Zhiyu and Yang, Hao and Qin, Luozheng and Gong, Jia and Yang, Mengping and Li, Hao},
  journal={arXiv preprint arXiv:2507.06119},
  year={2025}
}

@article{liao2025icve,
  title={In-context learning with unpaired clips for instruction-based video editing},
  author={Liao, Xinyao and Zeng, Xianfang and Song, Ziye and Fu, Zhoujie and Yu, Gang and Lin, Guosheng},
  journal={arXiv preprint arXiv:2510.14648},
  year={2025}
}

@article{decart2025lucyedit,
  title   = {Lucy Edit: Open-Weight Text-Guided Video Editing},
  author  = {DecartAI Team},
  year    = {2025}, 
  url     = { https://d2drjpuinn46lb.cloudfront.net/Lucy_Edit__High_Fidelity_Text_Guided_Video_Editing.pdf}
 }

@article{kimiteam2025kimilinear,
  title={Kimi linear: An expressive, efficient attention architecture},
  author={Team, Kimi and Zhang, Yu and Lin, Zongyu and Yao, Xingcheng and Hu, Jiaxi and Meng, Fanqing and Liu, Chengyin and Men, Xin and Yang, Songlin and Li, Zhiyuan and others},
  journal={arXiv preprint arXiv:2510.26692},
  year={2025}
}

@misc{qwen2026qwen35,
    title  = {{Qwen3.5}: Towards Native Multimodal Agents},
    author = {{Qwen Team}},
    month  = {February},
    year   = {2026},
    url    = {https://qwen.ai/blog?id=qwen3.5}
}

@article{wu2025qwenimagetechnicalreport,
  title={Qwen-image technical report},
  author={Wu, Chenfei and Li, Jiahao and Zhou, Jingren and Lin, Junyang and Gao, Kaiyuan and Yan, Kun and Yin, Sheng-ming and Bai, Shuai and Xu, Xiao and Chen, Yilei and others},
  journal={arXiv preprint arXiv:2508.02324},
  year={2025}
}

@article{yang2024gated,
  title={Gated delta networks: Improving mamba2 with delta rule},
  author={Yang, Songlin and Kautz, Jan and Hatamizadeh, Ali},
  journal={arXiv preprint arXiv:2412.06464},
  year={2024}
}

@article{yang2024gla,
  title={Gated linear attention transformers with hardware-efficient training},
  author={Yang, Songlin and Wang, Bailin and Shen, Yikang and Panda, Rameswar and Kim, Yoon},
  journal={arXiv preprint arXiv:2312.06635},
  year={2023}
}

@article{xie2024sana,
  title={Sana: Efficient high-resolution image synthesis with linear diffusion transformers},
  author={Xie, Enze and Chen, Junsong and Chen, Junyu and Cai, Han and Tang, Haotian and Lin, Yujun and Zhang, Zhekai and Li, Muyang and Zhu, Ligeng and Lu, Yao and others},
  journal={arXiv preprint arXiv:2410.10629},
  year={2024}
}

@article{su2021roformer,
  title={RoFormer: Enhanced Transformer with Rotary Position Embedding},
  author={Su, Jianlin and Lu, Yu and Pan, Shengfeng and Murtadha, Ahmed and Wen, Bo and Liu, Yunfeng},
  journal={arXiv preprint arXiv:2104.09864},
  year={2021}
}

@article{dao2024mamba2,
  title={Transformers are ssms: Generalized models and efficient algorithms through structured state space duality},
  author={Dao, Tri and Gu, Albert},
  journal={arXiv preprint arXiv:2405.21060},
  year={2024}
}

@article{zhang2019root,
  title={Root mean square layer normalization},
  author={Zhang, Biao and Sennrich, Rico},
  journal={Advances in neural information processing systems},
  volume={32},
  year={2019}
}

@inproceedings{schlag2021linear,
  title={Linear transformers are secretly fast weight programmers},
  author={Schlag, Imanol and Irie, Kazuki and Schmidhuber, J{\"u}rgen},
  booktitle={International conference on machine learning},
  pages={9355--9366},
  year={2021},
  organization={PMLR}
}

@article{sun2023retentive,
  title={Retentive network: A successor to transformer for large language models},
  author={Sun, Yutao and Dong, Li and Huang, Shaohan and Ma, Shuming and Xia, Yuqing and Xue, Jilong and Wang, Jianyong and Wei, Furu},
  journal={arXiv preprint arXiv:2307.08621},
  year={2023}
}

@article{xiao2024streamingllm,
  title={Efficient streaming language models with attention sinks},
  author={Xiao, Guangxuan and Tian, Yuandong and Chen, Beidi and Han, Song and Lewis, Mike},
  journal={arXiv preprint arXiv:2309.17453},
  year={2023}
}

@article{zhu2026causalforcing,
  title={Causal Forcing: Autoregressive Diffusion Distillation Done Right for High-Quality Real-Time Interactive Video Generation},
  author={Zhu, Hongzhou and Zhao, Min and He, Guande and Su, Hang and Li, Chongxuan and Zhu, Jun},
  journal={arXiv preprint arXiv:2602.02214},
  year={2026}
}

@article{huang2025selfforcing,
  title={Self forcing: Bridging the train-test gap in autoregressive video diffusion},
  author={Huang, Xun and Li, Zhengqi and He, Guande and Zhou, Mingyuan and Shechtman, Eli},
  journal={arXiv preprint arXiv:2506.08009},
  year={2025}
}

@article{cui2026lollongerlongerscaling,
  title={LoL: Longer than Longer, Scaling Video Generation to Hour},
  author={Cui, Justin and Wu, Jie and Li, Ming and Yang, Tao and Li, Xiaojie and Wang, Rui and Bai, Andrew and Ban, Yuanhao and Hsieh, Cho-Jui},
  journal={arXiv preprint arXiv:2601.16914},
  year={2026}
}

@article{chen2024diffusionforcing,
  title={Diffusion forcing: Next-token prediction meets full-sequence diffusion},
  author={Chen, Boyuan and Mart{\'\i} Mons{\'o}, Diego and Du, Yilun and Simchowitz, Max and Tedrake, Russ and Sitzmann, Vincent},
  journal={Advances in Neural Information Processing Systems},
  volume={37},
  pages={24081--24125},
  year={2024}
}

@article{wan2025wan,
  title={Wan: Open and advanced large-scale video generative models},
  author={Wan, Team and Wang, Ang and Ai, Baole and Wen, Bin and Mao, Chaojie and Xie, Chen-Wei and Chen, Di and Yu, Feiwu and Zhao, Haiming and Yang, Jianxiao and others},
  journal={arXiv preprint arXiv:2503.20314},
  year={2025}
}

@article{gao2025seedance,
  title={Seedance 1.0: Exploring the boundaries of video generation models},
  author={Gao, Yu and Guo, Haoyuan and Hoang, Tuyen and Huang, Weilin and Jiang, Lu and Kong, Fangyuan and Li, Huixia and Li, Jiashi and Li, Liang and Li, Xiaojie and others},
  journal={arXiv preprint arXiv:2506.09113},
  year={2025}
}

@inproceedings{charbon1994loss,
  title={Two deterministic half-quadratic regularization algorithms for computed imaging},
  author={Charbonnier, Pierre and Blanc-Feraud, Laure and Aubert, Gilles and Barlaud, Michel},
  booktitle={Proceedings of 1st international conference on image processing},
  volume={2},
  pages={168--172},
  year={1994},
  organization={IEEE}
}

@inproceedings{johnson2016perceptualloss,
  title={Perceptual losses for real-time style transfer and super-resolution},
  author={Johnson, Justin and Alahi, Alexandre and Fei-Fei, Li},
  booktitle={European conference on computer vision},
  pages={694--711},
  year={2016},
  organization={Springer}
}

@inproceedings{zou2025turbovaed,
  title={Turbo-vaed: Fast and stable transfer of video-vaes to mobile devices},
  author={Zou, Ya and Yao, Jingfeng and Yu, Siyuan and Zhang, Shuai and Liu, Wenyu and Wang, Xinggang},
  booktitle={Proceedings of the AAAI Conference on Artificial Intelligence},
  year={2026}
}

@inproceedings{zhang2025diffusion4k,
  title={Diffusion-4k: Ultra-high-resolution image synthesis with latent diffusion models},
  author={Zhang, Jinjin and Huang, Qiuyu and Liu, Junjie and Guo, Xiefan and Huang, Di},
  booktitle={Proceedings of the Computer Vision and Pattern Recognition Conference},
  pages={23464--23473},
  year={2025}
}

@article{ye2025ultraflux,
  title={UltraFlux: Data-Model Co-Design for High-quality Native 4K Text-to-Image Generation across Diverse Aspect Ratios},
  author={Ye, Tian and Fei, Song and Zhu, Lei},
  journal={arXiv preprint arXiv:2511.18050},
  year={2025}
}

@article{dao2022flashattention,
  title={Flashattention: Fast and memory-efficient exact attention with io-awareness},
  author={Dao, Tri and Fu, Dan and Ermon, Stefano and Rudra, Atri and R{\'e}, Christopher},
  journal={Advances in neural information processing systems},
  volume={35},
  pages={16344--16359},
  year={2022}
}

@article{dao2023flashattention2,
  title={Flashattention-2: Faster attention with better parallelism and work partitioning},
  author={Dao, Tri},
  journal={arXiv preprint arXiv:2307.08691},
  year={2023}
}

@article{hacohen2026ltx2,
  title={LTX-2: Efficient Joint Audio-Visual Foundation Model},
  author={HaCohen, Yoav and Brazowski, Benny and Chiprut, Nisan and Bitterman, Yaki and Kvochko, Andrew and Berkowitz, Avishai and Shalem, Daniel and Lifschitz, Daphna and Moshe, Dudu and Porat, Eitan and others},
  journal={arXiv preprint arXiv:2601.03233},
  year={2026}
}

@article{sla,
  title={Sla: Beyond sparsity in diffusion transformers via fine-tunable sparse-linear attention},
  author={Zhang, Jintao and Wang, Haoxu and Jiang, Kai and Yang, Shuo and Zheng, Kaiwen and Xi, Haocheng and Wang, Ziteng and Zhu, Hongzhou and Zhao, Min and Stoica, Ion and others},
  journal={arXiv preprint arXiv:2509.24006},
  year={2025}
}
\newpage

\appendix
\section{Mixed-Precision Quantization}
\label{appendix:mpq}

\noindent\textbf{Search setup and metrics.}
\emph{Quality.} For each candidate policy, we run the full 30-second streaming generation under the same input noise as the BF16 reference on 45 calibration prompts, and compare the produced output latent against the BF16 output latent. We report the relative RMSE and LPIPS averaged over the calibration set (lower is better), where RelRMSE normalizes the latent RMSE by the BF16 reference's standard deviation so that errors are comparable across prompts of different scale.
\emph{Compute accounting.} The DiT processes the 30-second video in chunks of 3 latent frames, so a single forward pass operates on a chunk of $3\times22\times40$ latent tokens. The FLOP totals in Table~\ref{tab:mpq_summary} are for this single per-chunk forward, computed as $\sum_\ell 2 \cdot \mathrm{in}_\ell \cdot \mathrm{out}_\ell \cdot N_\ell$ over every quantizable linear / point-wise convolution, with $N_\ell$ the tokens seen by layer $\ell$ ($N=2640$ for self-attention and FFN, $N=300$ for cross-attention key/value over text tokens). The full DiT then executes ${\sim}9.84$\,T FLOPs per chunk; reporting per-chunk FLOPs (rather than per-video) is sufficient because the FP4 / FP8 fractions and the resulting speedup factors are invariant to the chunk count.
\emph{Columns.} \textit{\#Params FP4} and \textit{FLOPs} are the parameters (M) and per-chunk FLOPs (G) demoted to FP4 in each configuration; the remaining linear layers stay in FP8. \textit{\%Param / \%FLOPs} are the corresponding shares of the full DiT. \textit{Speedup} is the idealized roofline factor over BF16, so all-FP8 gives 2x and all-FP4 gives 4x. To collapse quality and efficiency into a single scalar, we report \textit{Cost over Speedup} = $\mathrm{RMSE} / \mathrm{Speedup}$: the FP8 ratio of $0.23$ is the reference that guides us to identify the best mixed-precision policy.

\begin{table*}[h]
    \centering
    \caption{Mixed-precision search summary on the SANA hybrid DiT ($2$\,B parameters, $9.84$\,T FLOPs in a chunk generation of $3\times22\times40$ latent).}
    \label{tab:mpq_summary}
    \resizebox{\textwidth}{!}{%
    \begin{tabular}{lrrrrrrrr}
    \toprule
    \multirow{2}{*}{Config / module} & \#Params & FLOPs & \multirow{2}{*}{\%Param} & \multirow{2}{*}{\%FLOPs} & \multirow{2}{*}{RMSE $\downarrow$} & \multirow{2}{*}{LPIPS $\downarrow$} & \multirow{2}{*}{Speedup $\uparrow$} & Cost over  \\
    & FP4 (M) & (G) & & & & & & Speedup $\downarrow$ \\
    \midrule
    \multicolumn{9}{l}{\textit{Baselines}} \\
    FP8            & $0$    & $0$    & $0\%$    & $0\%$    & $0.45$ & $0.10$ & $2.00\times$ & $0.23$ \\
    FP4            & $2040$ & $9840$ & $100\%$  & $100\%$  & $1.04$ & $0.35$ & $4.00\times$ & $0.26$ \\
    \midrule
    \multicolumn{9}{l}{\textit{Single-group FP4 (only this group $\to$ FP4, rest FP8)}} \\
    FFN-Input                        & $602$  & $3179$ & $29\%$ & $32\%$ & $0.65$ & $0.17$ & $2.39\times$ & $0.27$ \\
    FFN-Output                       & $310$  & $1635$ & $15\%$ & $17\%$ & $0.57$ & $0.14$ & $2.18\times$ & $0.26$ \\
    FFN-Temporal                     & $310$  & $1635$ & $15\%$ & $17\%$ & $0.52$ & $0.12$ & $2.18\times$ & $0.24$ \\
    SA-O                             & $103$  & $545$  & $5\%$  & $6\%$  & $0.61$ & $0.15$ & $2.06\times$ & $0.30$ \\
    SA-Q                             & $103$  & $545$  & $5\%$  & $6\%$  & $0.47$ & $0.10$ & $2.06\times$ & $0.23$ \\
    SA-K                             & $103$  & $545$  & $5\%$  & $6\%$  & $0.49$ & $0.11$ & $2.06\times$ & $0.24$ \\
    SA-V                             & $103$  & $545$  & $5\%$  & $6\%$  & $0.56$ & $0.13$ & $2.06\times$ & $0.27$ \\
    CA-Q                             & $103$  & $545$  & $5\%$  & $6\%$  & $0.54$ & $0.12$ & $2.06\times$ & $0.26$ \\
    CA-O                             & $103$  & $545$  & $5\%$  & $6\%$  & $0.46$ & $0.10$ & $2.06\times$ & $0.22$ \\
    CA-KV                            & $201$  & $120$  & $10\%$ & $1\%$  & $0.49$ & $0.11$ & $2.01\times$ & $0.24$ \\
    \midrule
    \multicolumn{9}{l}{\textit{Block-range FP4 (all modules in this block range $\to$ FP4)}} \\
    Shallow Blocks (0--5)            & $612$  & $2950$ & $30\%$ & $30\%$ & $0.79$ & $0.22$ & $2.35\times$ & $0.34$ \\
    Middle Blocks (6--13)            & $816$  & $3940$ & $40\%$ & $40\%$ & $0.71$ & $0.19$ & $2.50\times$ & $0.29$ \\
    Deep Blocks (14--19)             & $612$  & $2950$ & $30\%$ & $30\%$ & $0.64$ & $0.16$ & $2.35\times$ & $0.27$ \\
    \midrule
    \multicolumn{9}{l}{\textit{Mixed-Precision}} \\
    Basic Mix (CA-O + SA-Q + SA-K + FFN-Temporal)              & $619$  & $3270$ & $30\%$ & $33\%$ & $0.56$ & $0.14$ & $2.40\times$ & $0.23$ \\
    Basic Mix + FFN-Input/Output (Deep blocks 14--19)          & $893$  & $4710$ & $44\%$ & $48\%$ & $0.60$ & $0.15$ & $2.63\times$ & $0.23$ \\
    \textbf{Basic Mix + FFN-Input/Output (Deep+Mid 6--19, Ours)} & $\mathbf{1258}$ & $\mathbf{6640}$ & $\mathbf{62\%}$ & $\mathbf{67\%}$ & $\mathbf{0.64}$ & $\mathbf{0.17}$ & $\mathbf{3.02\times}$ & $\mathbf{0.21}$ \\
    \bottomrule
    \end{tabular}%
    }
    \end{table*}

\noindent\textbf{Trade-off analysis.} We make the following observations from Table~\ref{tab:mpq_summary}.
\textbf{First}, the baselines reveal the central tension: pure FP4 has a higher cost over speedup than FP8 ($0.26$ vs.\ $0.23$), so doubling the raw compute speedup does not pay back the quality penalty. The goal of the search is therefore not to push as much as possible to FP4, but to find a mixed policy whose ratio is strictly below the FP8 reference of $0.23$.
\textbf{Second}, the single-group sweep spreads from $0.22$ (CA-O) to $0.30$ (SA-O), and only CA-O dips below the FP8 reference on its own; SA-Q, SA-K and FFN-Temporal sit just above, while the value/post-mixing projections (SA-V, SA-O, CA-Q) and FFN input/output project clearly above FP8, providing direct evidence for the per-layer decisions made above.
\textbf{Third}, holding the FP4 FLOP budget fixed at $30\%$ across the three block ranges flips the cost-over-speedup ratio from $0.34$ (shallow) through $0.29$ (middle) to $0.27$ (deep): the same FP4 budget hurts least when spent on the deeper blocks, motivating the depth-restricted demotion of FFN input/output.
\textbf{Finally}, the build-up in the mixed-precision section makes the trade-off concrete: Basic Mix matches the FP8 ratio ($0.23$ at $2.40\times$); adding FFN input/output in the deep blocks holds the ratio at $0.23$ while lifting the speedup to $2.63\times$; and only extending those FFN demotions into the middle blocks (\textit{Ours}) drops the ratio below FP8, reaching $0.21$ at an idealized $3.02\times$ speedup over BF16, dominating both baselines and every intermediate configuration we evaluated.

\begin{figure*}[t]
    \centering
    \begin{subfigure}[b]{0.48\textwidth}
        \centering
        \includegraphics[width=\textwidth]{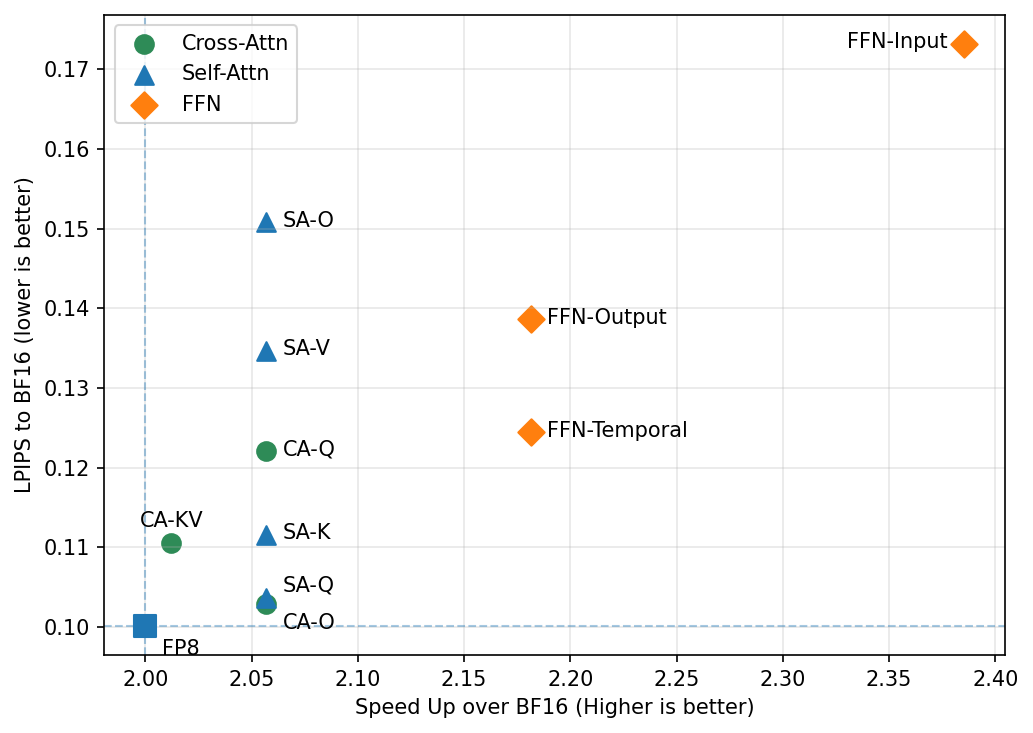} 
        \caption{Per-layer quantization policy search}
        \label{fig:mpq_lpips_layer}
    \end{subfigure}
    \hfill 
    \begin{subfigure}[b]{0.48\textwidth}
        \centering
        \includegraphics[width=\textwidth]{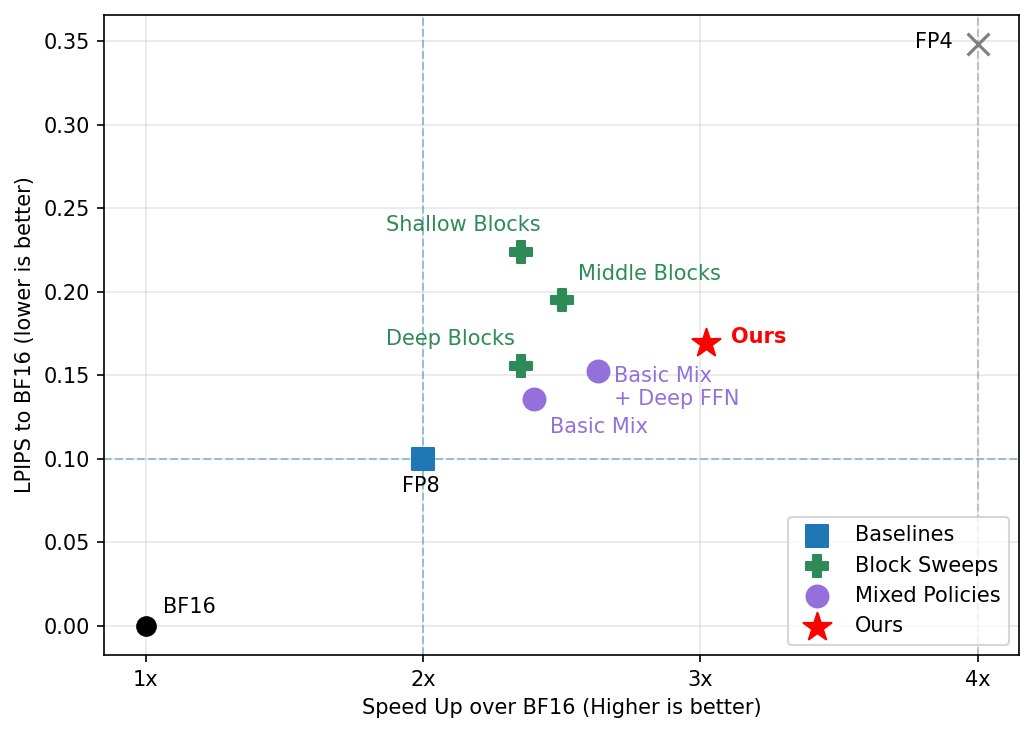} 
        \caption{Per-block and mixed quantization policy search}
        \label{fig:mpq_lpips_block}
    \end{subfigure}

    \caption{\textbf{Mixed-Precision Quantization Policy Search on LPIPS}. For each subfigure, the x-axis is the estimated speed up and the y-axis is the quantization error over the BF16 reference. The selected mixed-precision policy achieves the best trade-off between efficiency and quality.}
    \label{fig:mpq_lpips}
\end{figure*}

\section{Implementation Details}
\label{appendix:implementation}
\noindent\textbf{Model Design.}
\ours is a 2B hybrid diffusion transformer with the LTX2 VAE~\cite{hacohen2026ltx2} (compression ratio 32$\times$32$\times$8). The DiT contains 20 transformer blocks, including 5 evenly inserted softmax-attention blocks and 15 GDN blocks. The hidden size is 2240 with 20 attention heads.

\noindent\textbf{Dataset.}
We use the open-source Ditto~\cite{bai2025ditto} and OpenVE~\cite{he2025openve} datasets, as well as the human-centric editing data built by our data pipeline (Appendix~\ref{appendix:data}). The total number of short video clips is about 10M. For the pretraining stage, we filter the dataset based on our VLM verification results, with dataset-specific filtering criteria, obtaining 7M clips after filtering. In the SFT stage, we use stricter filtering criteria to obtain 1M clips. For the long-video dataset, we use an internal dataset with 10K one-minute videos and annotate the edit instruction and reverse instruction using our data pipeline.

\noindent\textbf{Model Training.}
The bidirectional model is trained with 32 NVIDIA H100 GPUs for about 100K iterations with batch size 2 per GPU. The learning rate is set to 5e-5 with the AdamW optimizer. For the long training stage, following LongLive~\cite{yang2025longlive}, we train the model in three stages: ODE initialization, self-forcing training, and streaming long training. The proposed cycle-reverse regularization is adopted in the long training stage.

\section{More Results}
\label{appendix:more-results}
\begin{figure*}
    \centering
    \includegraphics[width=\linewidth]{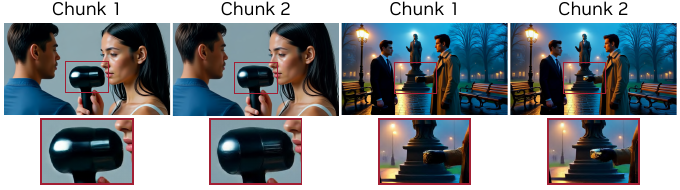}
    \caption{\textbf{Limitations of Linear Attention.} Purely linear attention~\cite{chen2025sanavideo} suffers from visible chunk-to-chunk appearance jumps and temporal flicker.}
    \label{fig:la-limitations}
\end{figure*}
\begin{figure}[t]
  \vspace{-3em}
  \centering
  \includegraphics[width=0.97\linewidth]{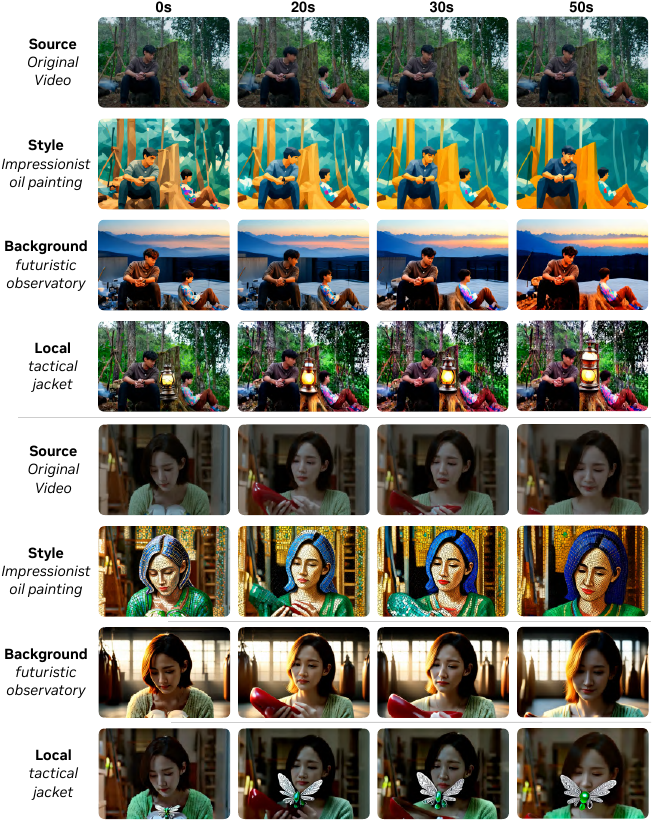}
  \vspace{-5pt}
  \caption{\textbf{More visualizations.} \ours follows editing instructions while preserving consistency with the original input videos.}
  \vspace{-1.5em}
  \label{fig:more-results}
\end{figure}

We present additional video editing results in Figure~\ref{fig:more-results}. For each group, the first row shows the original video as the source reference. The subsequent rows demonstrate our model's capability in three distinct editing tasks: style transfer (e.g., low poly or mosaic), background replacement (e.g., concrete terrace or boxing gym), and local object addition (e.g., adding a lantern or dragonfly). The results show that our method can precisely apply these modifications while maintaining high temporal consistency and structural fidelity to the original motion and character identity.
Figure~\ref{fig:la-limitations} further illustrates the chunk-boundary artifacts of pure linear attention, motivating the hybrid architecture used in \ours.

\section{GDN Kernel Details}
\label{appendix:kernel}

\noindent\textbf{Tensor shapes and layout.}
This appendix describes the Triton GDN kernel used in our
experiments. The input layout is $qkv\in\mathbb{R}^{B\times N\times 3\times
H\times D}$ with $N=F S$. In the production configuration, $D=112$ is padded to
\texttt{BLOCK\_D}=128 inside the kernels. The recurrent matrix state therefore
has shape $128\times128$ per $(B,H)$ stream, and the vector state has length
128. Intermediate frame summaries use shape $(B H,F,128,128)$ for matrix
terms and $(B H,F,128)$ for vector terms.

\begin{algorithm}[t]
\caption{Bidirectional GDN forward}
\label{alg:gdn-kernel}
\begin{algorithmic}[1]
\Require $qkv\in\mathbb{R}^{B\times FS\times 3\times H\times D}$,
         $\beta$, decay $\alpha$, normalization weights, RoPE tables
\Ensure output $O\in\mathbb{R}^{B\times FS\times H\times D}$
\State Pad head dimension to $\texttt{BLOCK\_D}=128$
\Comment{Phase A: frame-parallel summaries}
\ForAll{$(b,h,f)$ in parallel}
    \State $P_f \gets I - K_{\mathrm{rot},f}^{\top}\operatorname{diag}(\beta_f)K_{\mathrm{rot},f}$
    \State $A_f \gets K_{\mathrm{rot},f}^{\top}\operatorname{diag}(\beta_f)V_f$
    \State $P^z_f \gets I - K_f^{\top}\operatorname{diag}(\beta_f)K_f$
    \State $b_f \gets K_f^\top \beta_f$
\EndFor
\Comment{Phase B: compact recurrent scans}
\ForAll{$(b,h)$ in parallel}
    \State $S^{\rightarrow}_{-1},z^{\rightarrow}_{-1} \gets$ cached state or zero
    \For{$f=0$ to $F-1$}
        \State $S^{\rightarrow}_{f}\gets \alpha_f P_f S^{\rightarrow}_{f-1}+A_f$
        \State $z^{\rightarrow}_{f}\gets \alpha_f P^z_f z^{\rightarrow}_{f-1}+b_f$
    \EndFor
    \State $S^{\leftarrow}_{F-1},z^{\leftarrow}_{F-1}\gets 0$
    \For{$f=F-1$ down to $1$}
        \State $S^{\leftarrow}_{f-1}\gets \alpha_f P_f S^{\leftarrow}_{f}+A_f$
        \State $z^{\leftarrow}_{f-1}\gets \alpha_f P^z_f z^{\leftarrow}_{f}+b_f$
    \EndFor
    \For{$f=0$ to $F-1$}
        \State $S^{\mathrm{hist}}_f\gets S^{\rightarrow}_f+S^{\leftarrow}_f$
        \State $z^{\mathrm{hist}}_f\gets z^{\rightarrow}_f+z^{\leftarrow}_f$
    \EndFor
\EndFor
\Comment{Phase C: output streaming over spatial tiles}
\ForAll{$(b,h,f)$ in parallel}
    \For{$s_0=0$ to $S-1$ step $\texttt{BLOCK\_S}$}
        \State Load and normalize $Q_{f,s_0:s_0+\texttt{BLOCK\_S}}$
        \State Apply RoPE to obtain $Q_{\mathrm{rot}}$
        \State $N \gets Q_{\mathrm{rot}} S^{\mathrm{hist}}_f$
        \State $d \gets Q z^{\mathrm{hist}}_f$
        \State $O_{f,s_0:s_0+\texttt{BLOCK\_S}} \gets N/(d+\epsilon)$
    \EndFor
\EndFor
\end{algorithmic}
\end{algorithm}

\noindent\textbf{Implementation details.}
Phase A is implemented as two kernels: a matrix stream for $(P_f,A_f)$ and a
lighter vector stream for $(P^z_f,b_f)$. The vector stream does not need $V$ or
RoPE. Both streams accumulate frame summaries in fp32 and store $I-P$ directly,
so Phase B can compute $(I-P)S$ with one MMA. Phase B keeps the scan state in
fp32. In bidirectional mode it writes the reverse contribution directly into
the forward history buffer, so Phase C is launched once on the combined
history. Phase C then streams the spatial dimension of $Q_f$ in
\texttt{BLOCK\_S} tiles.

\noindent\textbf{Precision modes.}
The kernel supports three dot-product modes, inherited from the global GDN
precision setting:
\begin{itemize}
    \item \textbf{fp32}: Triton dot operands are fp32 with
    \texttt{input\_precision="ieee"}. The Phase-A to Phase-B HBM bridge is
    fp32, and Phase-C numerator/denominator buffers are fp32.
    \item \textbf{TF32}: the Phase-A to Phase-B bridge remains fp32, but
    matrix products use TF32 tensor cores.
    \item \textbf{bf16}: dot operands use bf16 tensor cores with fp32
    accumulation. Phase-A summaries and Phase-C output buffers are stored in
    bf16 to reduce HBM traffic, while the recurrent Phase-B scan state remains
    fp32.
\end{itemize}
For fp32 and TF32 modes, the frame summaries communicated between Phase A
and Phase B are stored in fp32. For bf16 mode, they are stored in bf16 to reduce
HBM traffic.

\noindent\textbf{Architecture-specific launch parameters.}
We tune Phase A, B, and C separately because they stress different resources.
Phase A and C are spatial streaming kernels controlled mainly by
\texttt{BLOCK\_S}. Phase B is a compact recurrent scan controlled mainly by
the number of warps and whether D-tiling is enabled. fp32 and TF32 share
the same Phase-A and Phase-C launch shapes, but Phase B is dispatched by the
exact dot-product mode. Table~\ref{tab:gdn-launch} shows the final effective
launch configuration used by the kernel.

\begin{table*}[t]
\centering
\setlength{\tabcolsep}{4pt}
\begin{tabular}{l l c c c}
\toprule
GPU & precision & Phase A $(nw,BS)$ & Phase B $(d,nw,ns,\mathrm{acc})$ & Phase C $(nw,BS)$ \\
\midrule
A100 & fp32 & $(16,32)$ & $(4,32,1,+)$ & $(16,32)$ \\
A100 & TF32       & $(16,32)$ & $(4,8,2,-)$  & $(16,32)$ \\
A100 & bf16       & $(8,32)$  & $(4,8,2,-)$  & $(4,32)$ \\
\midrule
H100 & fp32 & $(8,32)$  & $(4,32,1,+)$ & $(16,32)$ \\
H100 & TF32       & $(8,32)$  & $(4,8,2,-)$  & $(16,32)$ \\
H100 & bf16       & $(8,64)$  & $(4,8,2,-)$  & $(8,32)$ \\
\midrule
GB200 & fp32 & $(8,128)$ & $(8,4,1,-)$ & $(4,64)$ \\
GB200 & TF32       & $(8,128)$ & $(4,8,2,-)$ & $(4,64)$ \\
GB200 & bf16       & $(4,64)$  & $(4,8,2,-)$ & $(8,64)$ \\
\midrule
5090 & fp32 & $(8,16)$ & $(8,4,1,-)$ & $(8,16)$ \\
5090 & TF32       & $(8,16)$ & $(8,8,1,-)$ & $(8,16)$ \\
5090 & bf16       & $(8,32)$ & $(8,8,1,-)$ & $(4,32)$ \\
\midrule
GB10 & fp32 & $(8,16)$ & $(8,4,1,-)$ & $(8,16)$ \\
GB10 & TF32       & $(8,16)$ & $(1,8,1,-)$ & $(8,16)$ \\
GB10 & bf16       & $(8,32)$ & $(1,8,1,-)$ & $(4,32)$ \\
\bottomrule
\end{tabular}
\caption{Final Triton launch parameters. Across phases, $nw$ is the number of
warps, $ns$ is the number of pipeline stages, and $BS$ is the spatial block
size. For Phase B, $d$ is the number of output-column splits ($d=1$ means no
D-tiling), and $\mathrm{acc}$ indicates whether accumulator fusion is enabled.}
\label{tab:gdn-launch}
\end{table*}

\noindent\textbf{D-tiled Phase B.}
For Phase B, the output columns of the matrix recurrence are independent:
\[
S_f[:,J] = \alpha_f P_f S_{f-1}[:,J] + A_f[:,J].
\]
We exploit this by splitting the output-column dimension into $d_{\mathrm{splits}}$
tiles. Each Triton program owns one column tile $J$, which lowers live state
size and increases grid parallelism. The vector state $z_f$ is not
column-splittable, so only the leading D-tile computes it. Because D-tiling
also duplicates reads of $P_f$, the selected policy depends on GPU and
precision, as shown in Table~\ref{tab:gdn-launch}.

\noindent\textbf{Benchmark.}
We evaluate three RoPE/cache strategies: fixed-RoPE, rolling-RoPE with sink,
and rolling-RoPE. All runs use the same output chunk pattern, with an initial
five-frame chunk followed by three-frame chunks. In
Table~\ref{tab:gdn-headline-speedups}, \emph{total} is the end-to-end sampling
time, \emph{ch0} is the first GDN call that emits five frames, and \emph{ch1+}
is the median subsequent GDN call that emits three frames. Depending on the
cache strategy, a GDN call may also consume cached recurrent state or replay
cached pre-RoPE context internally. We report two measurements. First, the
end-to-end sampling time is the wall-clock time of the diffusion
sampling phase; video preparation, VAE decoding, encoding, and saving are
excluded. Second, per-call GDN time is measured inside the model with CUDA
events around each GDN forward call. Calls are bucketed by strategy and chunk
type, and we report the median time for the initial chunk
(\emph{ch0}) and the subsequent chunks (\emph{ch1+}). For each
configuration we run PyTorch and the Triton GDN kernel under the same precision
setting and compute speedup as the ratio of PyTorch time to Triton-kernel time.
The reported sampling time uses the second generated sample to avoid
one-time warmup effects. Table~\ref{tab:gdn-headline-speedups} reports
PyTorch and Triton-kernel runtimes, with speedups over PyTorch in parentheses
on the Triton rows. In fp32, per-call GDN speedups are
$1.4\times$--$13.8\times$ across five GPU
classes and three rope strategies; end-to-end sampling speedups are
$1.42\times$--$2.65\times$ because non-GDN work remains in the pipeline. At
bf16, per-call speedups are $2.3\times$--$10.6\times$ and sampling speedups
are $1.28\times$--$2.01\times$.

\begin{table*}[t]
\centering
\footnotesize
\setlength{\tabcolsep}{1.5pt}
\newcommand{\gdntime}[2]{\begin{tabular}{@{}c@{}}#1\\($#2\times$)\end{tabular}}
\begin{tabular}{c c c c c c c c c c c c}
\toprule
& & & \multicolumn{3}{c}{fixed} & \multicolumn{3}{c}{sink} & \multicolumn{3}{c}{rolling} \\
\cmidrule(lr){4-6}\cmidrule(lr){7-9}\cmidrule(lr){10-12}
GPU & precision & impl. & total (s) & ch0 (ms) & ch1+ (ms) & total (s) & ch0 (ms) & ch1+ (ms) & total (s) & ch0 (ms) & ch1+ (ms) \\
\midrule
\multirow[t]{4}{*}{A100} & \multirow[t]{2}{*}{fp32} & PyTorch & 232.69 & 72.30 & 67.15 & 237.52 & 75.77 & 69.98 & 252.31 & 72.78 & 79.45 \\
& & Kernel & \gdntime{87.72}{2.65} & \gdntime{7.20}{10.04} & \gdntime{4.87}{13.78} & \gdntime{94.26}{2.52} & \gdntime{13.04}{5.81} & \gdntime{10.47}{6.68} & \gdntime{99.30}{2.54} & \gdntime{12.96}{5.61} & \gdntime{14.37}{5.53} \\
& \multirow[t]{2}{*}{bf16} & PyTorch & 57.50 & 12.07 & 7.85 & 60.63 & 12.08 & 8.85 & 69.51 & 12.32 & 13.90 \\
& & Kernel & \gdntime{43.03}{1.34} & \gdntime{3.45}{3.49} & \gdntime{2.47}{3.18} & \gdntime{46.07}{1.32} & \gdntime{4.02}{3.01} & \gdntime{3.74}{2.36} & \gdntime{48.40}{1.44} & \gdntime{3.99}{3.08} & \gdntime{4.99}{2.78} \\
\midrule
\multirow[t]{4}{*}{H100} & \multirow[t]{2}{*}{fp32} & PyTorch & 32.35 & 9.35 & 5.99 & 35.02 & 9.33 & 7.53 & 43.88 & 9.61 & 13.20 \\
& & Kernel & \gdntime{21.69}{1.49} & \gdntime{6.06}{1.54} & \gdntime{2.88}{2.08} & \gdntime{24.62}{1.42} & \gdntime{5.78}{1.61} & \gdntime{5.24}{1.44} & \gdntime{27.72}{1.58} & \gdntime{3.84}{2.50} & \gdntime{5.58}{2.37} \\
& \multirow[t]{2}{*}{bf16} & PyTorch & 28.86 & 7.23 & 4.73 & 30.89 & 7.18 & 5.26 & 36.36 & 7.40 & 8.43 \\
& & Kernel & \gdntime{18.88}{1.53} & \gdntime{1.58}{4.56} & \gdntime{1.05}{4.49} & \gdntime{20.86}{1.48} & \gdntime{1.94}{3.71} & \gdntime{1.71}{3.09} & \gdntime{22.19}{1.64} & \gdntime{1.92}{3.85} & \gdntime{2.61}{3.23} \\
\midrule
\multirow[t]{4}{*}{GB200} & \multirow[t]{2}{*}{fp32} & PyTorch & 21.65 & 6.44 & 4.54 & 23.30 & 6.34 & 5.27 & 30.46 & 6.70 & 9.96 \\
& & Kernel & \gdntime{13.71}{1.58} & \gdntime{2.40}{2.68} & \gdntime{1.55}{2.93} & \gdntime{15.40}{1.51} & \gdntime{2.50}{2.54} & \gdntime{2.53}{2.08} & \gdntime{16.94}{1.80} & \gdntime{2.53}{2.65} & \gdntime{3.51}{2.84} \\
& \multirow[t]{2}{*}{bf16} & PyTorch & 22.32 & 6.21 & 4.55 & 23.51 & 6.19 & 5.37 & 30.84 & 6.23 & 9.12 \\
& & Kernel & \gdntime{13.65}{1.64} & \gdntime{1.41}{4.42} & \gdntime{1.51}{3.02} & \gdntime{15.52}{1.51} & \gdntime{1.54}{4.02} & \gdntime{2.33}{2.30} & \gdntime{15.43}{2.00} & \gdntime{1.48}{4.21} & \gdntime{2.22}{4.10} \\
\midrule
\multirow[t]{4}{*}{5090} & \multirow[t]{2}{*}{fp32} & PyTorch & 89.28 & 11.40 & 24.74 & 92.59 & 11.43 & 26.86 & 103.43 & 11.72 & 33.21 \\
& & Kernel & \gdntime{49.16}{1.82} & \gdntime{5.08}{2.24} & \gdntime{2.95}{8.39} & \gdntime{51.73}{1.79} & \gdntime{5.30}{2.16} & \gdntime{4.54}{5.92} & \gdntime{53.13}{1.95} & \gdntime{4.23}{2.77} & \gdntime{4.98}{6.67} \\
& \multirow[t]{2}{*}{bf16} & PyTorch & 46.03 & 9.39 & 5.72 & 47.17 & 9.08 & 6.55 & 55.17 & 9.56 & 10.51 \\
& & Kernel & \gdntime{35.83}{1.28} & \gdntime{3.20}{2.94} & \gdntime{2.11}{2.71} & \gdntime{36.59}{1.29} & \gdntime{3.48}{2.61} & \gdntime{2.60}{2.52} & \gdntime{39.72}{1.39} & \gdntime{3.59}{2.66} & \gdntime{3.74}{2.81} \\
\midrule
\multirow[t]{4}{*}{GB10} & \multirow[t]{2}{*}{fp32} & PyTorch & 397.58 & 117.57 & 93.70 & 425.10 & 117.94 & 107.12 & 491.71 & 120.16 & 149.10 \\
& & Kernel & \gdntime{193.76}{2.05} & \gdntime{16.55}{7.10} & \gdntime{10.02}{9.35} & \gdntime{212.71}{2.00} & \gdntime{18.25}{6.46} & \gdntime{18.11}{5.92} & \gdntime{229.11}{2.15} & \gdntime{18.34}{6.55} & \gdntime{26.87}{5.55} \\
& \multirow[t]{2}{*}{bf16} & PyTorch & 368.64 & 101.09 & 83.62 & 393.57 & 101.44 & 91.01 & 434.69 & 102.39 & 116.46 \\
& & Kernel & \gdntime{186.15}{1.98} & \gdntime{12.48}{8.10} & \gdntime{7.88}{10.62} & \gdntime{206.15}{1.91} & \gdntime{15.83}{6.41} & \gdntime{13.47}{6.75} & \gdntime{216.68}{2.01} & \gdntime{15.77}{6.49} & \gdntime{19.81}{5.88} \\
\bottomrule
\end{tabular}
\caption{GDN-kernel runtimes and speedups over PyTorch. PyTorch rows report
baseline runtimes; Kernel rows report Triton-kernel runtimes with speedup in
parentheses. Column headers give units: \emph{total} is end-to-end sampling
time in seconds, while \emph{ch0} and \emph{ch1+} are per-call GDN times in
milliseconds for the first chunk and later chunks, respectively. The three
RoPE/cache strategies are fixed-RoPE, rolling-RoPE with sink, and rolling-RoPE.}
\label{tab:gdn-headline-speedups}
\end{table*}

\section{Data Pipeline}
\label{appendix:data}
Our data pipeline aims to construct high-quality human-centric video editing pairs, enabling controllable semantic edits and temporally coherent motion.
In the following subsections, we present the construction process for both short and long videos, along with the corresponding quality control procedures.
Figure~\ref{fig:data_pipeline} summarizes the complete pipeline, including short-video pair construction, long-video instruction construction, and VLM-based filtering.

\subsection{Short Video}
\noindent\textbf{Edit Instruction Generation and First Frame Edit.} We first generate edit instructions with a taxonomy-guided prompting strategy. For each source video, we sample an editing task from four high-level categories: local human edits, background edits, artistic style transfer, and composite edits.
\textit{Local edits} include outfit replacement, outfit color or pattern changes, hairstyle changes, hair color changes, and accessory edits.
\textit{Background edits} either replace the scene while preserving the subject, or jointly adapt the background and outfit to form a coherent contextual transformation.
\textit{Style-transfer} instructions are sampled from a diverse style dictionary, while \textit{composite edits} combine multiple attributes such as fashion, hair, makeup, setting, season, decade, or character archetype.
Given the sampled task, we prompt a vision-language model, Qwen3VL~\cite{qwen3technicalreport}, to analyze the source video and produce a single self-contained editing instruction with concrete visual details such as material, color, silhouette, texture, lighting interaction, and style coherence.
We then apply this edit to the first frame of the video. Specifically, we use Qwen-Image-Edit~\cite{wu2025qwenimagetechnicalreport} conditioned on the edit instruction, the first frame, and the pose extracted from the first frame to obtain an edited first frame, which serves as a visual anchor for later edit-video generation.

\noindent\textbf{In-Context Generation for Edit Video.}
To enable fine-grained and controllable video editing, we provide the edited first frame and the source video to Qwen3VL~\cite{qwen3technicalreport} to generate a caption describing the desired target video content. Meanwhile, we extract a pose video from the source video to serve as a motion anchor, preserving temporal dynamics. Conditioned on these signals, we generate the edited video using a controllable text-image-video-to-video pipeline, Wan2.2-Fun-Control~\cite{aigc_apps_VideoX_Fun_2026}. For each sample, the generated caption serves as the text prompt, the pose video provides motion guidance, and the edited first frame is used as both the initial frame and a visual reference.

\subsection{Long Video}
To support streaming long training and cycle-reverse regularization, we construct an additional long-video subset with paired forward and backward edit instructions.
For each long source video, we first sample a representative anchor frame, typically the first frame, and use it as the visual evidence for instruction generation.
Conditioned on the anchor frame, we use Gemini-3-Flash to generate the forward edit instruction $p^{+}$.
The edit prompt is sampled from five edit families: background replacement, local addition, local removal, local replacement, and style transfer, following the taxonomy-based approach described above.

We construct the backward instruction $p^{-}$ using the source video and its forward instruction $p^{+}$.
The VLM is asked to infer the inverse operation that would recover the observed source content after the forward edit has been applied.

\subsection{Data Verification}
\noindent\textbf{VLM Verification for Short Video.}
After generating candidate edited videos, we apply VLM-based verification to filter low-quality samples.
The verifier compares each edited video against the original source video, the edit instruction, and the predicted edited-video caption.
This comparison is necessary because a generated video can be visually plausible while still failing to follow the intended edit, altering non-edit regions, or introducing temporal artifacts.

We evaluate each candidate along four dimensions.
\textit{Instruction alignment} measures whether the edited video faithfully implements the requested semantic change.
\textit{Non-edit consistency and temporal stability} measures whether regions not targeted by the edit are preserved from the source video and whether the output remains stable across frames.
\textit{Physical plausibility} evaluates lighting consistency, geometry, motion, and other real-world constraints after editing.
\textit{Video quality} captures sharpness, color stability, compression artifacts, and overall visual coherence.
Each dimension is scored on a 0--10 scale, and the verifier also returns brief comments describing strengths and failure cases.
The resulting scores and textual comments are saved for each sample and used to select the final training data.

\noindent\textbf{Data Verification and VLM Verification for Long Videos.}
Long videos may contain title cards or black-screen segments that provide little useful editing content.
We sample the first and last 10 seconds at approximately 2 FPS, and discard videos with excessive near-black pixel ratio.
Moreover, for long-video self-forcing distillation, the student is bounded by the teacher's editing capability; if the teacher cannot execute a forward edit, the corresponding long sample provides little useful supervision.
We therefore verify long-video instructions by taking the first 81 frames of each source video and asking the teacher model to generate the edited 81-frame clip conditioned on the forward instruction.
A VLM then compares the source clip, the teacher-edited clip, and the instruction, deciding whether the specified edit is successfully performed. If not, we discard the sample.

\section{Causal VAE Training}
\label{appendix:vae}

We train only the decoder of the causal VAE, while keeping the pretrained VAE encoder fixed.
Given a training video $x \in \mathbb{R}^{B \times C \times T \times H \times W}$, the frozen encoder produces a latent code $z$, and the causal decoder reconstructs $\hat{x}=D_{\theta}^{\mathrm{causal}}(z)$.
The bidirectional LTX2 decoder is used as a frozen teacher for intermediate feature supervision.
Below we expand the loss terms in Eq.~\ref{eq:causal-vae-objective}.
For compact notation, $x_{b,t}$ and $\hat{x}_{b,t}$ denote the $t$-th frame of the target and reconstruction in video $b$.

\noindent\textbf{Charbonnier reconstruction loss.}
For robust pixel-level reconstruction, we use a Charbonnier penalty on the decoded video:
\begin{equation}
    \mathcal{L}_{\mathrm{charb}}
    =
    \frac{1}{BCTHW}
    \sum_{b,c,t,h,w}
    \sqrt{
        \left(\hat{x}_{b,c,t,h,w} - x_{b,c,t,h,w}\right)^2
        + \epsilon^2
    },
    \qquad \epsilon=10^{-6}.
\end{equation}
This term behaves similarly to an $\ell_1$ reconstruction loss for large residuals while remaining smooth around zero, which stabilizes decoder-only fine-tuning.

\noindent\textbf{Perceptual loss.}
To preserve perceptual similarity beyond per-pixel accuracy, we apply the AlexNet-based LPIPS loss independently to each video frame:
\begin{equation}
    \mathcal{L}_{\mathrm{perc}}
    =
    \frac{1}{BT}
    \sum_{b,t}
    \mathcal{D}_{\mathrm{LPIPS}}^{\mathrm{Alex}}\left(\hat{x}_{b,t}, x_{b,t}\right).
\end{equation}

\noindent\textbf{Haar wavelet loss.}
The Haar wavelet term explicitly supervises high-frequency details.
Let $\mathcal{W}_{\mathrm{H}}(\cdot)$ denote a single-level 2D Haar transform applied frame by frame, and let
$\mathcal{W}_{\mathrm{H}}^{\mathrm{high}}(\cdot)$ concatenate the three high-frequency subbands LH, HL, and HH, omitting the low-frequency LL band.
We minimize
\begin{equation}
    \mathcal{L}_{\mathrm{haar}}
    =
    \frac{1}{BT}
    \sum_{b,t}
    \frac{1}{M_{\mathrm{H}}}
    \left\|
    \mathcal{W}_{\mathrm{H}}^{\mathrm{high}}\left(\hat{x}_{b,t}\right)
    -
    \mathcal{W}_{\mathrm{H}}^{\mathrm{high}}\left(x_{b,t}\right)
    \right\|_{1}.
\end{equation}
Here $M_{\mathrm{H}}$ is the number of high-frequency Haar coefficients per frame.
This loss encourages the causal decoder to recover sharp edges and fine texture that are often smoothed by purely pixel-level objectives.

\noindent\textbf{Intermediate feature distillation.}
We further align the causal student decoder with the bidirectional teacher decoder using intermediate activations.
Let $F_{\ell}^{S}(z)$ and $F_{\ell}^{T}(z)$ be the student and teacher decoder features at layer $\ell$, where $\ell$ is selected from the decoder mid-block and upsampling blocks.
The teacher is frozen and gradients are stopped through its features:
\begin{equation}
    \mathcal{L}_{\mathrm{distill}}
    =
    \sum_{\ell \in \mathcal{S}}
    \frac{1}{N_{\ell}}
    \left\|
        F_{\ell}^{S}(z)
        -
        F_{\ell}^{T}(z)
    \right\|_{2}^{2},
\end{equation}
where $\mathcal{S}$ is the set of distilled decoder blocks and $N_{\ell}$ is the number of elements in the feature tensor.

\section{Frame-wise Gated DeltaNet}
\label{appendix:gdn}

This section provides the detailed formulation of the frame-wise Gated DeltaNet (GDN) block used in \ours, which extends the Gated Delta Network~\cite{yang2024gated} to the frame-wise streaming video setting. The design goal of this block is to provide a compact, finite recurrent memory for causal streaming generation: instead of caching raw key/value tokens from all previous chunks, each GDN block only carries forward the terminal recurrent states from the previous chunk, so the cache size is independent of the number of streamed chunks.

\subsection{Notation}
All equations below are presented for a single attention head; the full multi-head computation runs $H$ heads in parallel and concatenates their outputs along the channel dimension. For a chunk of $F$ frames with $N$ spatial tokens per frame, the per-head latent feature is
\[
    x \in \mathbb{R}^{F \times N \times D},
\]
where $D$ is the per-head channel dimension. For the $f$-th frame, the per-head input tokens are
\[
    x_f \in \mathbb{R}^{N \times D}.
\]
For convenience we use the column-token convention in the following derivation,
\[
    X_f = x_f^\top \in \mathbb{R}^{D \times N},
\]
and we omit the head index $h$ throughout, with the understanding that all weight matrices are head-specific. Each head maintains two recurrent states: a KV memory
\[
    S^{kv}_f \in \mathbb{R}^{D \times D}
\]
and a normalizer state
\[
    S^z_f \in \mathbb{R}^{D}.
\]

\subsection{Feature parameterization}
Given $X_f$, the block computes per-head query, key, and value features by pointwise (kernel size $1$) linear projections:
\begin{equation}
    Q_f = W_q X_f, \qquad
    K_f = W_k X_f, \qquad
    V_f = W_v X_f,
\end{equation}
where $W_q, W_k, W_v \in \mathbb{R}^{D \times D}$ and $Q_f, K_f, V_f \in \mathbb{R}^{D \times N}$. We then apply RMSNorm \cite{zhang2019root}, a ReLU kernel feature map following Sana's linear attention design \cite{xie2024sana}, and a fixed 3D RoPE \cite{su2021roformer} to the query and key:
\begin{equation}
    \hat{Q}_f
    = \rho_{\mathrm{3D}}\!\Bigl(\,
        \mathrm{ReLU}\bigl(\mathrm{RMSNorm}(Q_f)\bigr)
    \Bigr),
    \qquad
    \hat{K}_f
    = \rho_{\mathrm{3D}}\!\Bigl(\,
        \mathrm{ReLU}\bigl(\mathrm{RMSNorm}(K_f)\bigr)
    \Bigr).
\end{equation}
To keep the magnitude of the normalizer projection $K_f^\top S^z_f$ stable for large spatial token counts, both $K_f$ and $\hat{K}_f$ are additionally rescaled by $1/\sqrt{D N}$ before entering the recurrence. The unrotated key $K_f$ is used for the normalizer state, while the rotated key $\hat{K}_f$ is used for the KV memory update and readout; this ``RoPE-on-numerator-only'' choice preserves mass conservation in the linear-attention denominator, since rotated $\hat K_f$ does not sum to a positive scalar after frame-wise accumulation.

\subsection{Gate parameterization}
The block predicts three gates: a frame-wise decay gate $\alpha_f$, a token-wise write gate $\beta_f$, and an output gate $G_f$.

\noindent\textbf{Decay gate $\alpha_f$.} We adopt a Mamba-style discretized state-space parameterization \cite{dao2024mamba2} for the decay gate, which empirically yields substantially more stable long-horizon recurrence than a plain sigmoid:
\begin{equation}
    \bar{x}_f = \frac{1}{N} \sum_{n=1}^{N} \bigl(X_f\bigr)_{:, n}
    \;\in\; \mathbb{R}^{D},
\end{equation}
\begin{equation}
    \Delta_f = \mathrm{softplus}\!\bigl(W_\alpha\, \bar{x}_f + b_\alpha\bigr)
    \;\in\; \mathbb{R}_{>0},
    \qquad
    \alpha_f = \exp\!\bigl(-\,e^{A_{\log}} \cdot \Delta_f\bigr)
    \;\in\; (0, 1],
    \label{eq:appendix_gdn_alpha}
\end{equation}
where $W_\alpha \in \mathbb{R}^{1 \times D}$ projects the spatially pooled frame feature $\bar{x}_f$ to a scalar, $b_\alpha \in \mathbb{R}$ is a bias, and $A_{\log} \in \mathbb{R}$ is a learnable per-head log-rate parameter that ensures $A = e^{A_{\log}} > 0$. The decay $\alpha_f$ is therefore a per-frame per-head scalar shared across all $N$ spatial positions and all $D$ channels of the head. This parameterization is equivalent to the discretization $\alpha = e^{-A \Delta}$ of a continuous-time linear ODE with data-dependent step size $\Delta$, and matches the selective-scan gating of Mamba-2.

\noindent\textbf{Write gate $\beta_f$.} The write gate is a token-wise sigmoid:
\begin{equation}
    \beta_f = \mathrm{sigmoid}\bigl(W_\beta X_f + b_\beta\bigr)
    \;\in\; (0, 1)^{1 \times N},
    \label{eq:appendix_gdn_beta}
\end{equation}
where $W_\beta \in \mathbb{R}^{1 \times D}$ produces a single scalar per token (per head). For convenience we write
\[
    B_f = \mathrm{Diag}(\beta_f) \in \mathbb{R}^{N \times N}
\]
for its diagonal-matrix form, used in the matrix update equations below. The write gate controls how aggressively each token's delta-rule correction is written into memory.

\noindent\textbf{Output gate $G_f$.} The output gate is a per-token, per-channel SiLU-modulated linear projection:
\begin{equation}
    G_f = \mathrm{SiLU}\bigl(W_g X_f + b_g\bigr)
    \;\in\; \mathbb{R}^{D \times N},
    \label{eq:appendix_gdn_g}
\end{equation}
where $W_g \in \mathbb{R}^{D \times D}$. This is analogous to the output gate of Gated Linear Attention \cite{yang2024gla} and the gated MLP of Mamba and serves as a fine-grained channel-wise selector on the attention output.

\subsection{Frame-wise delta-rule update}
At frame $f$, the previous recurrent states are first decayed:
\begin{equation}
    \widetilde{S}^{kv}_f = \alpha_f\, S^{kv}_{f-1},
    \qquad
    \widetilde{S}^{z}_f  = \alpha_f\, S^{z}_{f-1}.
\end{equation}
The KV memory is then updated by a delta-rule correction in the spirit of fast-weight programmers \cite{schlag2021linear,yang2024gated}:
\begin{equation}
    S^{kv}_f
    =
    \widetilde{S}^{kv}_f
    +
    \bigl(
        V_f - \widetilde{S}^{kv}_f \hat{K}_f
    \bigr)
    B_f
    \hat{K}_f^{\!\top}.
    \label{eq:appendix_gdn_kv_update}
\end{equation}
Equivalently, this can be written in the compact delta-rule form
\begin{equation}
    S^{kv}_f
    =
    \alpha_f\, S^{kv}_{f-1}
    \bigl(
        I - \hat{K}_f B_f \hat{K}_f^{\!\top}
    \bigr)
    +
    V_f B_f \hat{K}_f^{\!\top}.
    \label{eq:appendix_gdn_kv_compact}
\end{equation}
The normalizer state is updated in the same correction-based manner, using the unrotated key $K_f$ to preserve mass conservation:
\begin{equation}
    S^z_f
    =
    \widetilde{S}^{z}_f
    +
    K_f B_f
    \bigl(
        \mathbf{1} - K_f^{\!\top}\, \widetilde{S}^{z}_f
    \bigr),
    \label{eq:appendix_gdn_z_update}
\end{equation}
where $\mathbf{1}\in\mathbb{R}^{N}$ is an all-one vector. Equivalently,
\begin{equation}
    S^z_f
    =
    \alpha_f
    \bigl(
        I - K_f B_f K_f^{\!\top}
    \bigr)
    S^z_{f-1}
    +
    K_f B_f \mathbf{1}.
    \label{eq:appendix_gdn_z_compact}
\end{equation}
The key difference from simple gated accumulation \cite{yang2024gla,sun2023retentive} lies in the residual term
\[
    V_f - \widetilde{S}^{kv}_f \hat{K}_f .
\]
Instead of directly writing $V_f \hat{K}_f^{\!\top}$ into memory, GDN first \emph{predicts} the current values from the decayed recurrent state and only writes back the remaining error, scaled by $\beta_f$. This correction-based update reduces interference between stored key/value associations \cite{schlag2021linear}, providing a more controlled finite-state memory for long streaming sequences. It is critical for minute-long streaming video generation where the same recurrent state must accumulate consistent identity and scene information across thousands of frames.

\subsection{Normalized readout and output projection}
After the recurrent states are updated, the output is obtained by a normalized linear-attention readout:
\begin{equation}
    Y_f
    =
    \frac{
        S^{kv}_f \hat{Q}_f
    }{
        (S^z_f)^{\!\top} Q_f \;+\; \epsilon
    },
    \label{eq:appendix_gdn_readout}
\end{equation}
where the numerator $S^{kv}_f \hat{Q}_f \in \mathbb{R}^{D \times N}$ uses the rotated query, the denominator $(S^z_f)^{\!\top} Q_f \in \mathbb{R}^{1 \times N}$ uses the unrotated query and is broadcast over the $D$ channel dimension, and $\epsilon = 10^{-6}$ is a small constant for numerical stability. The final per-head output is produced by an element-wise multiplication with the output gate followed by a linear projection,
\begin{equation}
    O_f
    =
    W_o\bigl(\, G_f \odot Y_f \,\bigr),
    \label{eq:appendix_gdn_output}
\end{equation}
where $W_o \in \mathbb{R}^{D \times D}$ and $\odot$ denotes the Hadamard product. Per-head outputs $\{O_f^{(h)}\}_{h=1}^{H}$ are then concatenated along the channel dimension and passed through a final shared linear projection.

\subsection{Streaming cache}
During streaming inference, each GDN block only caches the terminal states
\[
    \bigl(S^{kv},\; S^z\bigr) \;\in\; \mathbb{R}^{D \times D} \times \mathbb{R}^{D}
\]
per head from the previous chunk, and uses them to initialize the next chunk. The total cache size is therefore $\mathcal{O}\!\bigl(H D^2\bigr)$ per layer and is independent of the number of streamed chunks or the total streaming length. This allows the GDN blocks in \ours to serve as a compact \emph{global accumulated memory} that compresses the entire streaming history into a fixed-size matrix state, while the softmax blocks complement them by performing local window-and-sink refinement \cite{xiao2024streamingllm} over recent tokens.
\section{Safeguards}
\label{appendix:safeguards}

For deployment, the model
is paired with safeguards outside the generative model itself. We use a
layered design in which policy checks are applied before generation, during
generation when needed, and after generation before content is returned.

\noindent\textbf{Input screening.}
User prompts and conditioning inputs is first checked by a policy
classifier, such as Llama Guard~\cite{inan2023llama} or an equivalent safety classifier. The
classifier flags requests involving disallowed sexual content, minors,
graphic violence, self-harm, instructions for wrongdoing, privacy violations,
and other categories defined by the deployment policy. Prompts that clearly
violate policy are rejected. Ambiguous prompts can be routed to a stricter
policy, rewritten into a safe form, or sent for human review.

\noindent\textbf{Generation-time controls.}
The generation service should preserve the original safety decision throughout
sampling. For example, the system can attach a policy state to each request and
avoid later prompt expansions that introduce disallowed entities or actions.
Rate limits, authentication, and abuse monitoring should be applied to reduce
automated misuse. For high-risk categories, the service can use conservative
decoding settings or disable generation entirely rather than relying only on
post-hoc filtering.

\noindent\textbf{Output screening.}
Generated videos are checked before release. A practical implementation
can sample frames from the beginning, middle, and end of each video, run an
image or vision-language safety classifier on those frames, and optionally run
OCR to detect unsafe or private text rendered in the video. If any sampled
frame is flagged, the system blocks the video or route it to manual
review. For applications with stricter requirements, all frames or short
temporal clips can be screened instead of sparse frame samples.

\noindent\textbf{Provenance and traceability.}
Generated content includes provenance metadata or a watermark when the
deployment setting supports it. The service retains minimal audit logs
needed for abuse investigation, including the safety decision, model version,
and policy version, while avoiding unnecessary storage of sensitive user data.

\noindent\textbf{Evaluation.}
Safeguards will be evaluated with a held-out red-team set covering direct
policy violations, paraphrases, multilingual prompts, prompt-injection attempts,
and benign prompts that are easy to over-block. We report both block rates on
unsafe prompts and false-positive rates on benign prompts. Because safety
classifiers are imperfect, the safeguard layer should be treated as a risk
reduction mechanism rather than a proof that unsafe outputs cannot occur.

\section{Broader Impacts}
\label{appendix:broader-impacts}
SANA-Streaming enables real-time, high-quality video editing, which may benefit applications such as content creation, live broadcasting, and human–computer interaction. At the same time, such technology could be misused to generate misleading or manipulated video content, including real-time deepfakes or deceptive visual edits.

To mitigate these risks, our pipeline incorporates several safeguards,
including VLM-based data verification to filter low-quality or misaligned samples, as well as instruction consistency and temporal stability checks during data construction.
These measures aim to improve the faithfulness and reliability of generated content. However, we acknowledge that such safeguards are not sufficient to fully prevent misuse or unintended outcomes, especially in open or real-world deployment scenarios. Careful consideration of responsible deployment will be important to ensure that the technology is used in a socially beneficial manner.


\clearpage

\end{document}